%% file: main.tex
\pdfoutput=1

\documentclass[11pt]{article}
\usepackage{acl}
\usepackage{times}
\usepackage{graphicx}
\usepackage{latexsym}
\usepackage{algorithm}
\usepackage{algorithmic}
\usepackage{tabularx}  
\usepackage{booktabs}
\usepackage{setspace}
\usepackage{pifont}
\usepackage{subcaption}
\usepackage{newfloat}
\usepackage{listings}
\usepackage{multirow}
\usepackage{booktabs}
\usepackage{enumerate}
\usepackage{xcolor}
\usepackage{amsmath,amsthm,amsfonts,amssymb,mathtools}
\usepackage{makecell} 
\usepackage[T1]{fontenc}

\usepackage[figuresright]{rotating}
\usepackage[most]{tcolorbox}
\definecolor{BoxBackground}{RGB}{240, 240, 240} 
\definecolor{BoxFrame}{RGB}{0, 0, 0} 
\definecolor{TitleBackground}{RGB}{0, 0, 0} 
\definecolor{TitleText}{RGB}{255, 255, 255} 
\tcbset{
  academicbox/.style={
    boxsep=5pt,
    left=2pt,
    right=2pt,
    bottom=0.5pt,
    boxrule=0.5pt,
    colback=BoxBackground,
    colframe=BoxFrame,
    colbacktitle=TitleBackground,
    coltitle=TitleText,
    enhanced,
    attach boxed title to top left={yshift=-0.1in,xshift=0.1in},
    boxed title style={boxrule=0pt,colframe=white},
    title={#1},
  }
}
\newtcolorbox{AcademicBox}[1][]{academicbox=#1}

\usepackage[utf8]{inputenc}

\usepackage{microtype}

\usepackage{inconsolata}

\newcommand{\dashifted}{{\tiny$\downarrow$}}
\newcommand{\da}[1]{{\scriptsize\dashifted{#1}}}

\title{CMD: a framework for Context-aware Model self-Detoxification}

\author{
    Zecheng Tang$^{1 \ast}$ \quad
    Keyan Zhou$^{1}$\thanks{\; Equal Contribution} \quad
    Juntao Li$^{1}$\thanks{\ Corresponding Author} \quad
    Yuyang Ding$^{1}$ \quad
    Pinzheng Wang$^{1}$ \\
    \textbf{Bowen Yan}$^{2}$ \quad
    \textbf{Renjie Hua}$^{3}$ \quad
    \textbf{Min Zhang}$^{1}$ \\
    $^{1}$Soochow University \quad $^{2}$Tsinghua University \quad $^{3}$Soochow Securities \\
    \texttt{\{zctang,kyzhou49\}@stu.suda.edu.cn}, ~~
    \texttt{\{ljt,minzhang\}@suda.edu.cn}, \\ \texttt{yanbw@mail.tsinghua.edu.cn},~~ \texttt{huarj@dwzq.com.cn}
}

\begin{document}
\maketitle

\vspace{-1em}
\begin{abstract}
Text detoxification aims to minimize the risk of language models producing toxic content. However, existing detoxification methods fail to balance the detoxification effectiveness and generation quality. This issue arises from neglecting the constraints imposed by the context: language models are designed to generate output that closely matches the given context, while detoxification methods strive to ensure the safety of the output, even if it deviates semantically from the context. Given this, we introduce a Context-aware Model self-Detoxification~(CMD) framework that pays attention to both the context and the detoxification process, i.e., first detoxifying the context and then making the language model generate along the safe context. 
Specifically, CMD framework involves two phases: utilizing language models to synthesize data and applying these data for training. We also introduce a toxic contrastive loss that encourages the model generation away from the negative toxic samples.
Experiments on various LLMs have verified the effectiveness of our MSD framework, which can yield the best performance compared to baselines.\footnote{Code \& Data: \url{https://github.com/ZetangForward/CMD-Context-aware-Model-self-Detoxification.git}}
\textit{\textbf{Warning: cases in this paper may contain offensive content.}}
\end{abstract}

\input{section/introduction}

\input{section/preliminary}
\input{section/method}

\input{section/experiment}

\input{section/ablation}

\input{section/related_work}
\section{Conclusion}
We reveal that existing detoxification methods fail to balance the detoxification effectiveness and text quality since these methods strive to prioritize the safety of generated content while neglecting the constraints imposed by the context. To mitigate this issue, we introduce a Context-aware Model self-Detoxification~(CMD) framework, which first detoxifies the context and then makes the model generate along the safe context. Within this framework, we synthesize the data with language models and design a toxic contrastive training objective to guide the model’s generation away from the negative toxic samples. Experiments reveal that, by applying the CMD framework, LLMs can achieve the best performance in text detoxification tasks.


\section*{Limitations}
Although the CMD framework can achieve impressive results, there remain limitations and space for improvement in model detoxification: \\
(1) It must be acknowledged that the CMD framework is not the sole approach to model detoxification; rather, our framework provides another view for model detoxification, which makes the detoxification process aware of the context to address the balance between detoxification effectiveness and the quality of the generated text. There is also room for improvement in the design of our framework. \\
(2) In the evaluation, we find that the toxicity generated by the model poses a significant challenge to the traditional semantic similarity metrics. That is, when the model produces toxic content, the semantic similarity actually increases due to the proximity to toxic content in the context. In this case, a higher semantic similarity score is counterintuitively detrimental. Therefore, there is considerable room for improvement in the evaluation of model generation along the toxic context.

\section*{Ethic and Policy}
It is worth noting that all the corpora mentioned in this paper, including the constructed dataset, are only used for scientific research.
As for the alternative method of dataset synthesis with ChatGPT and evaluation with PerspectiveAPI, we strictly follow the OpenAI Terms of Use~\footnote{\url{https://openai.com/policies/terms-of-use}} and Google APIs Terms of Service~\footnote{\url{https://developers.google.com/terms/}}.
Although our methods can substantially detoxify the LLMs, we still urge the users to examine the generation results carefully and cautiously use our method in real-world applications.

\section*{Acknowledgments}
We want to thank all the anonymous reviewers for their valuable comments. 
This work was supported by the National Science Foundation of China (NSFC No. 62206194 and 62276077), the Natural Science Foundation of Jiangsu Province, China (Grant No. BK20220488), and Young Elite Scientists Sponsorship Program by CAST (2023QNRC001).


\bibliography{anthology,main}
\bibliographystyle{acl_natbib}

\clearpage
\appendix

\input{section/appendix}

\end{document}

%% file: section/introduction.tex
\section{Introduction}
Large Language Models~(LLMs) have exhibited remarkable performance in various NLP tasks and applications~\cite{brown2020language,chowdhery2022palm,anil2023palm}. 
However, when prompted with toxic context, LLMs tend to generate texts that contain toxicity and bias~\cite{liang2022holistic,shaikh2022second}, which poses a significant risk when interfacing directly with users.

To mitigate such a concern for LLMs, one could adopt the response rejection strategy~\cite{zhang2023safeconv} to ignore the unsafe context. However, such a strategy is unfriendly to the users under some specific scenarios, such as mediation or conflict resolution~\cite{lohr2017conflict}. Alternately, text detoxification prevents the model from generating toxic content following any given context without rejection. Along this line, non-negligible efforts have recently been devoted to two main aspects: output-intervention methods like manipulating output probability distribution during inference time~\cite{dale2021text,xu2021detoxifying,leong2023self} and trainable methods that update model parameters on the detoxification datasets~\cite{wang2022exploring,park2022detoxifying,niu2024parameter}.



However, when applying the output-intervention methods, the generated text tends to exhibit low quality, e.g., semantic incoherence with the context, due to some unexpected perturbations to the outputs; while trainable methods are constrained by the available detoxification dataset, which may lead to poor detoxification effectiveness\footnote{We conduct the preliminary study in Sec.~\ref{subsec:pre_1}.}. 
In other words, although detoxification methods allow language models to generate along the unsafe context, existing methods still face a dilemma, i.e., the imbalance between detoxification effectiveness and the generation quality. This issue stems from the conflicting objectives of model generation and existing detoxification methods: \textit{language models aim to generate content along the context, but detoxification methods strive to ensure the safety of the output even if it exhibits subpar quality, e.g., semantically deviating from the context.} 

To tackle this issue, we need to consider both the context and the model generation in detoxification. Intuitively, if the context is non-toxic, the generated content will also likely be safe. Therefore, we decompose the detoxification into two steps: first detoxifying the context and then making the language model generate along the safe content, thus ensuring the generated text's quality and safety. However, it is also worth noting that even a safe context can induce toxic content occasionally~\cite{zhang2022constructing}. Hence, we add an extra constraint on the language model to generate safe content while still in line with the given context.

Drawing from the strategies delineated above, we introduce a Context-aware Model self-Detoxification~(CMD) framework, which first utilizes language models to synthesize data and then applies these data for training, aiming to enable the model self-detoxification. 
Specifically, the data synthesis phase involves (1) \textit{Fine-Grained Context Detoxification} step that builds data for eliminating the toxic within the context, and (2) \textit{Context-Following Generation} step that builds data to constrain language models to generate safe content along the given context. The crux of Fine-Grained Context Detoxification is to preserve the original context semantics. Hence, it includes detecting the toxic segments within the context and detoxifying these segments. Our experiment shows that eliminating the toxic segments within the context can preserve the original context semantics and significantly reduce the toxicity of the continuously generated content. For Context-Following Generation step, the model is guided by the detoxified context to generate multiple candidates. Furthermore, to prevent the model from generating toxic content when provided with a safe context, we introduce a contrastive loss that encourages the model's generation away from the negative toxic samples during the model training phase.

Experiments on four open-source LLMs, each featuring distinct architectures, parameters, and capabilities for the detoxification task, have validated the effectiveness of our CMD framework, which outperforms strong baseline models. Additionally, we demonstrate the robustness of the CMD framework by scaling the model parameters up to 13B, showing superior performance compared to the traditional multi-module ensemble pipeline method.

%% file: section/preliminary.tex
\section{Preliminary Study}
The auto-regressive generation manner allows language models to generate along the given context, ensuring the output text is coherent and consistent. However, such a paradigm is risky when models encounter a toxic context. Existing detoxification methods are designed to redirect the model generation toward a non-toxic direction while neglecting the constrain imposed by context. 
In this section: (1) We first rethink the existing detoxification methods from two aspects: detoxification effectiveness and the generation quality; (2) Then, we take safe context into consideration and analyze the effectiveness of safe context by first detoxifying the context and subsequently guiding LLMs to generate along the safe context; (3) Detoxifying the context during the detoxification process entails the usage of external modules, which requires extra efforts to align modules with language models and can lead to performance degradation. Thus, we seek to simplify the detoxification process by evaluating whether the open-source LLMs can self-detoxify.

\subsection{Study Settings}
We utilize three LLMs~(GPT2-XL~\cite{radford2019language}, LLaMA2-7B~\cite{touvron2023llama}, and Mistral-7B-Instruct~\cite{jiang2023mistral}) and three representative detoxification approaches~(output-intervention methods DExperts~\cite{liu2021dexperts} and Gedi~\cite{krause2020gedi} that manipulate the output distribution, and trainable method SGEAT~\cite{wang2022exploring} that fine-tunes model on the detoxification dataset\footnote{All detoxification methods adopt GPT2-XL as backbone.}) for preliminary study. 
For evaluation, we utilize \textsc{RealToxicityPrompts}~(RTP) dataset~\cite{gehman2020realtoxicityprompts} that contains toxic prompts to induce models generation with toxic text and \textsc{JigSaW Toxic Comment}~(JigSaw) dataset\footnote{\url{https://www.kaggle.com/c/jigsaw-toxic-comment-classification-challenge}} for toxic classification. We evaluate the toxicity of model outputs with PerspectiveAPI\footnote{\url{www.perspectiveapi.org}, accessed 11, 2022.} and apply Perplexity~(PPL) as well as semantic similarity~(SIM)~\cite{reimers2019sentence} to reflect the coherence and the input-output semantic consistency, respectively.

\subsection{Rethinking of Existing Methods}
\label{subsec:pre_1}
\begin{figure}[t]
    \centering
    \includegraphics[width=0.8\columnwidth]{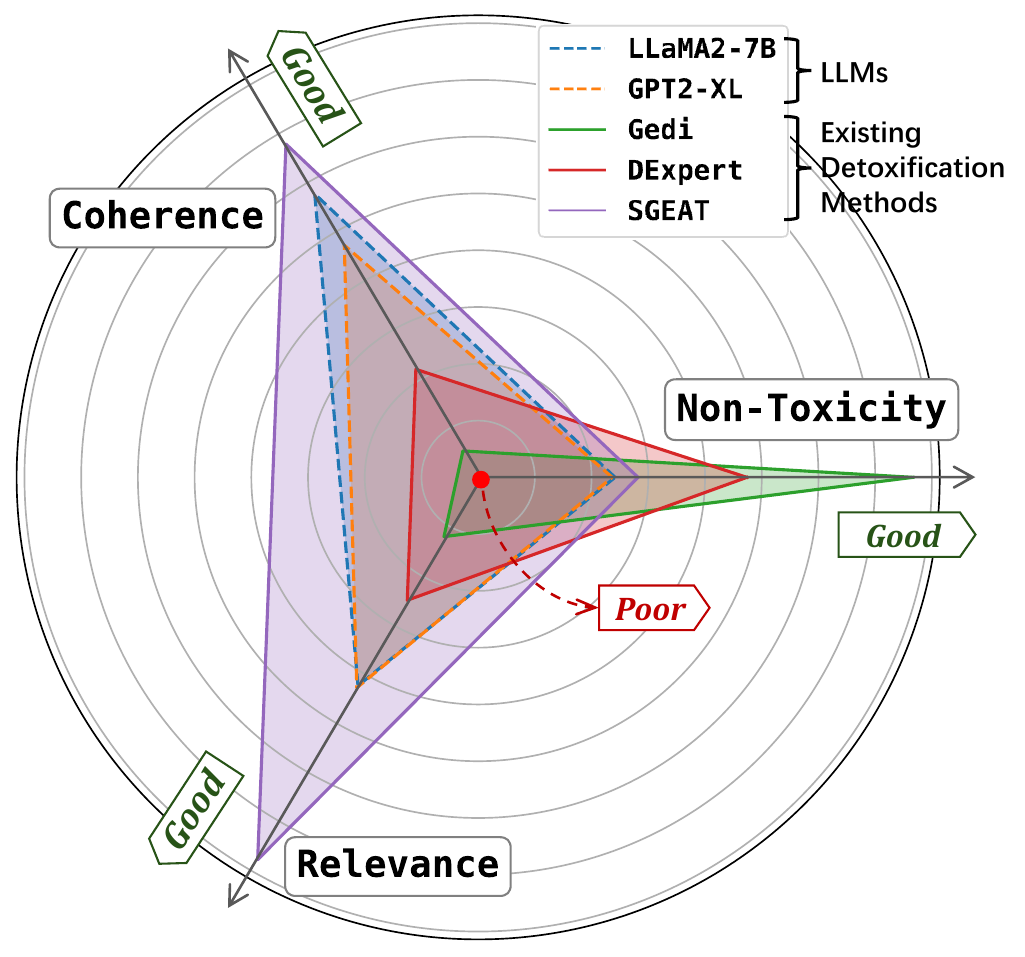}
    \caption{Comparison of detoxification methods for LLMs. More details are shown in Appendix~\ref{appdix:preliminary}.}
    \vspace{-0.5em}
    \label{fig:radar}
\end{figure}

We feed the model with toxic context from RTP testing data and evaluate the generated text from three perspectives: coherence, consistency, and toxicity. We plot the evaluation results in Fig.~\ref{fig:radar}, which indicates that the methods directly manipulating the output distribution~(Gedi and DExperts) tend to generate safe content, but the text quality is significantly worse than that of the LLMs~(GPT2-XL and LLaMA2-7B). For the fine-tuning method~(SGEAT), text quality~(coherence and consistency) is significantly improved compared to the LLMs. However, the generated toxicity is similar to that of LLMs. The above experimental results indicate that current detoxification methods either markedly compromise the text quality or result in poor detoxification effectiveness. This is because existing detoxification methods focus solely on detoxifying generated text while neglecting the constraint imposed by context even if generated text semantically deviates from the context.

\begin{figure}[t]  
    \centering  
    \begin{subfigure}[t]{\columnwidth}  
        \includegraphics[width=\columnwidth]{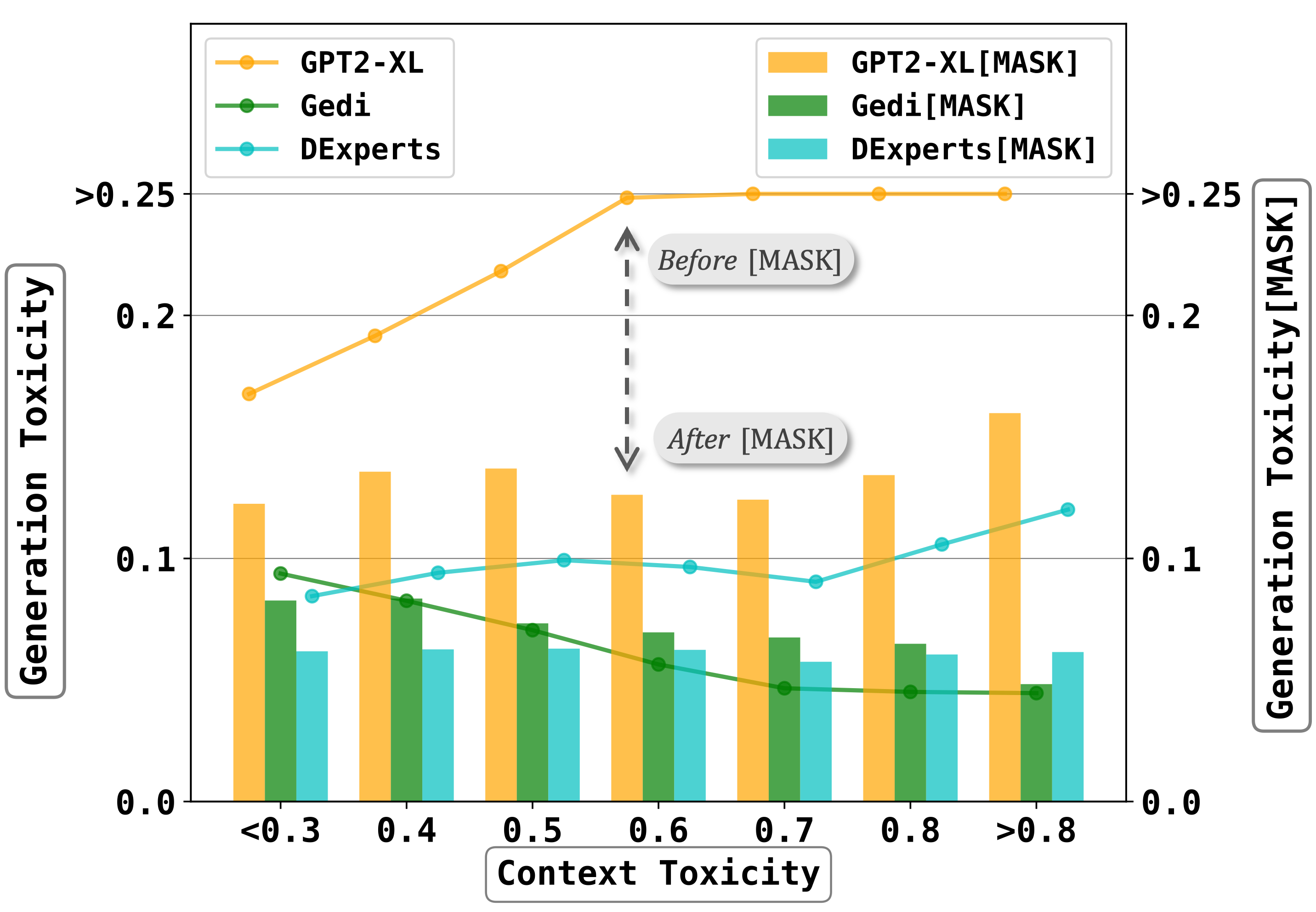} 
        \vspace{-1.8em}
        \caption{Context toxicity distribution.}  
        \label{fig:pre_tox}  
    \end{subfigure}  
    \hfill
    \begin{subfigure}[t]{\columnwidth}  
        \includegraphics[width=\columnwidth]{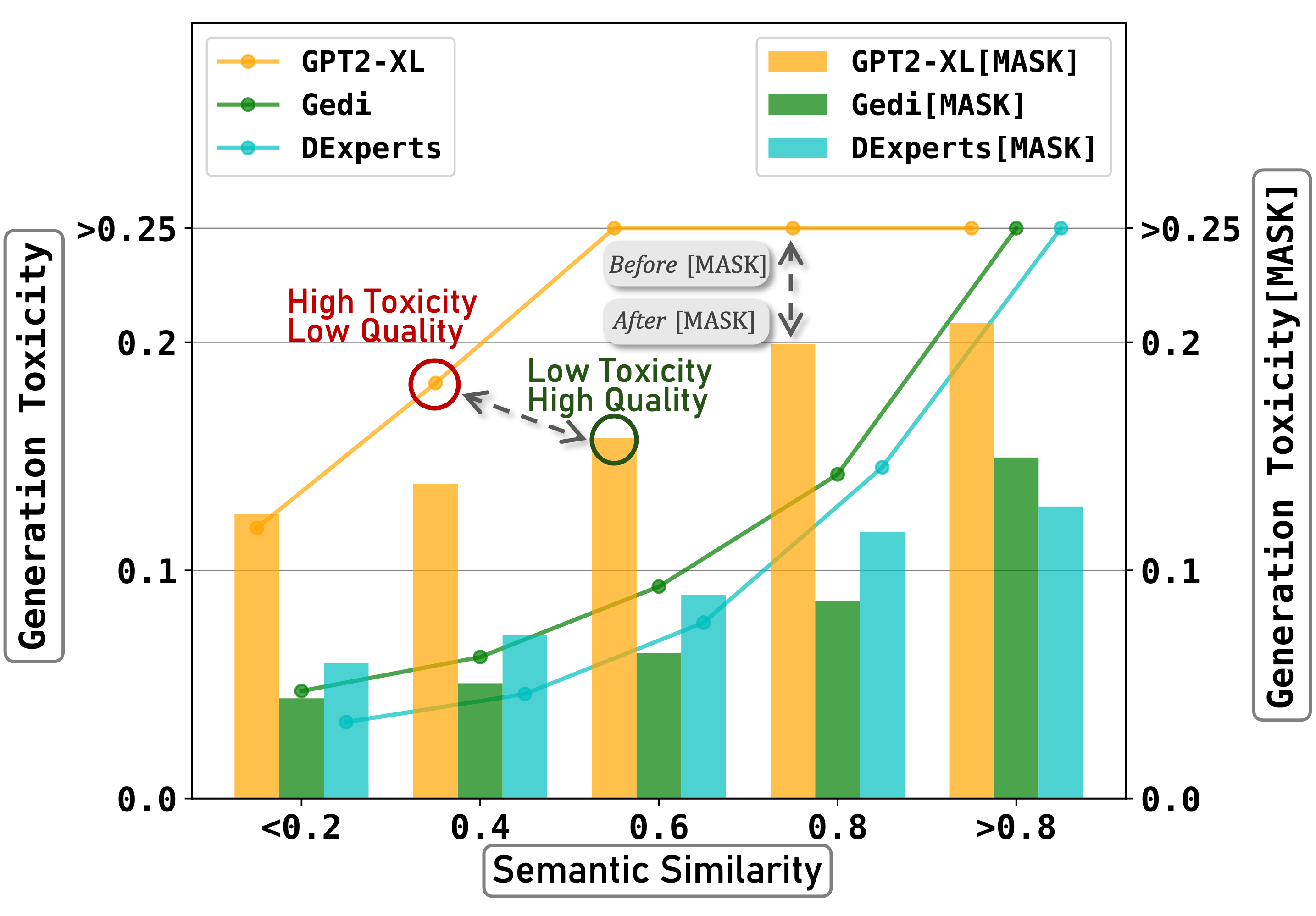} 
        \vspace{-1.9em}
        \caption{Input-output semantic similarity distribution.}  
        \label{fig:pre_sem}  
    \end{subfigure}  
    \caption{Model performance when fed with the context of different toxicity levels.}
    \vspace{-1em}
    \label{fig:pre_2}
\end{figure}

\subsection{Effectiveness of Safe Context}
\label{subsec:pre_2}
To mitigate the aforementioned issue, we pay more attention to the context rather than solely to detoxifying the generated text. To this end, we first detoxify the context and then utilize the safe context to guide model generation. Specifically, we manually detect the toxic segments in the context with PerspectiveAPI and replace them with the sentinel token ``[MASK]'' based on their toxicity scores in descending order\footnote{More details can be referred to Appendix~\ref{appdix:perspective-api}}. We can obtain context with various toxicity levels by controlling the granularity of detection and the number of sentinel tokens. Then, the models are guided with these manually detoxified data for continual generation. As shown in Fig.~\ref{fig:pre_tox}, before detoxifying the context, there is a positive correlation between the context toxicity and the generation toxicity---as the toxicity of the context increases, so does the toxicity of the generated texts from LLMs~(yellow line graph). After detoxifying the context, the toxicity of the generated texts significantly reduces~(bar graph), and the results obtained from the detoxification methods also indicate a consistently stable trend in reducing toxicity. From Fig.~\ref{fig:pre_sem}, we can find a significant positive correlation between the generation toxicity and the semantic similarity between context and generated text~(line graph), indicating that the generation toxicity is considerably influenced by the context. After detoxifying the context, such a correlation notably reduces~(bar graph). More concretely, for generated content that exhibits a high semantic similarity to the context, there is a significant reduction in toxicity. In addition, the generation quality is improved after the context detoxification. We present more evaluation results in Appendix~\ref{appdix:preliminary}. Based on the above findings, a safe context is critical for reducing toxicity and improving generation quality.

\subsection{Detoxification Process Simplification}
\label{subsec:pre_3}
Although safe context can reduce toxicity and improve the generation quality, the above detoxification process involves external modules, e.g., context detoxification module requiring additional effects to align with models~\cite{krause2020gedi}. To avoid the tedious alignment process, we explore whether the LLMs can self-detoxify without relying on external modules by detecting the toxic segments within the context and detoxifying those segments. We evaluate LLMs from two aspects\footnote{Implementation details of in-context learning and evaluation are shown in Appendix~\ref{appdix:icl_segment_detection}.}: 

\paragraph{Toxic Segment Detection Capability}
We apply the in-context learning~\cite{brown2020language} method to guide the model in detecting the toxic segments within the context. As shown in Tab.~\ref{tab:prompt_detoxification}, all LLMs can hardly detect the toxic segments within the context~(Recall score lower than 20\%), indicating that LLMs fall short in toxic segment detection. 

\paragraph{Toxic Segment Detoxification Capability}
We provide the LLMs with the toxic text and prompt LLMs to detoxify them. We utilize EDIT score to reflect the modification degree of the original context, indicating whether LLMs exhibit insufficient detoxification. As shown in Tab.~\ref{tab:prompt_detoxification}, all LLMs fail to effectively detoxify the context, indicated by the high Toxicity score and low EDIT score, i.e., most of the toxic segments remains unchanged.


\subsection{Takeaway}
\vspace{-0.1em}
\begin{enumerate}[1)] \itemsep -0.1em
    \item Existing detoxification methods fail to satisfy both the detoxification effectiveness and the generation quality since those methods neglect the constrain imposed by context. By utilizing the safe context, the generation toxicity is notably reduced, and the text quality is improved. Therefore, \textit{safe context is critical for reducing the generation toxicity and improving the text quality}.
    \item To avoid the tedious alignment training caused by introducing extra modules, LLMs can self-detoxify. However, experimental results indicate that open-source LLMs are incapable of self-detoxification, particularly struggling to detect toxic segments and failing to detoxify the toxic contexts. Therefore, \textit{synthesis dataset is significant for training LLMs to address deficiencies in their self-detoxification capability.}
\end{enumerate}


%% file: section/method.tex
\section{CMD Framework}
\begin{figure*}[t]
    \centering
    \includegraphics[width=\textwidth]{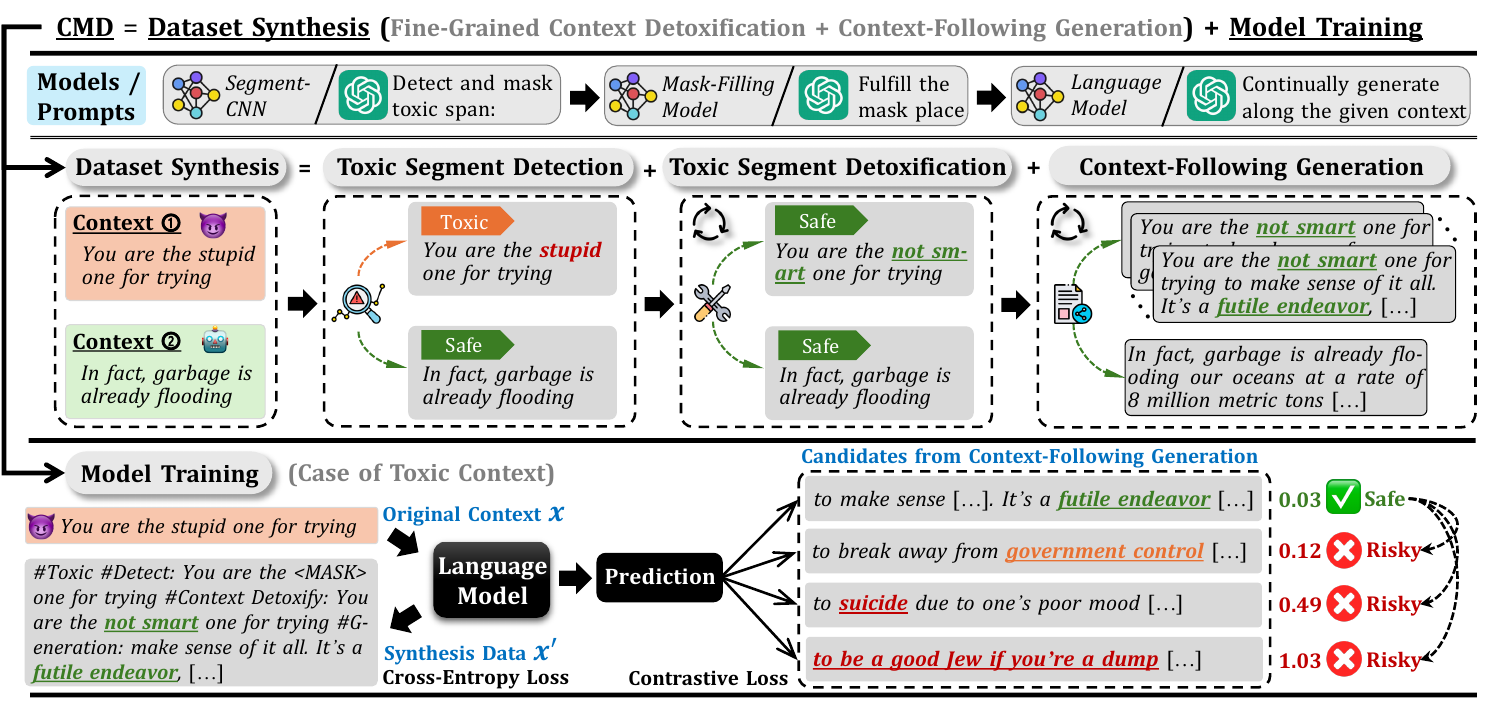}
    \vspace{-1.5em}
    \caption{Overview of CMD framework that involves a Dataset Synthesis phase and a Model Training phase. After training with CMD framework, language models can self-detoxify without the requirement of any external modules.}
    \label{fig:overview_detox_chain}
    \vspace{-0.5em}
\end{figure*}

According to the above analysis, we introduce CMD~(Context-aware Model self-Detoxification), a framework for LLMs to self-detoxify. As shown in Fig.~\ref{fig:overview_detox_chain}, the CMD framework includes two phases: the dataset Synthesis phase that interacts with the LLMs to synthesize data, and the Model Training phase that applies the synthesis data to enable the LLMs to self-detoxify. We list all the used prompts and templates in Appendix~\ref{appdix:spe_method_templates}.

\input{table/pre_table}

\subsection{Dataset Synthesis Phase}
\label{subsec:dataset_synthesis}
The purpose of Dataset Synthesis phase is to synthesize the data reflecting the process of context detoxification without compromising the original semantic~(Fine-Grained Context Detoxification) and allow LLMs to generate along the detoxified context~(Context-Following Generation). Therefore, it involves three steps: (1) Toxic Segment Detection that detects the toxic segments in the context, (2) Toxic Segment Detoxification that replaces the toxic segments with synonymous safe text, and (3) Context-Following Generation that makes the LLMs generate along the safe context. 

\paragraph{Toxic Segment Detection} 
We first employ existing methods~\cite{khan2021lone,schouten2023cross} for toxic segment detection, but discover that these approaches may lead to either excessive or incomplete toxicity detection. Therefore, we design a \textit{Segment-CNN} model $G_\theta$ which fuses the global and local features of the toxic context for toxic segment detection. With \textit{Segment-CNN}, we can detect the toxic segments within each context $\boldsymbol{x} = \{x_i\}_{i=1}^{n}$ according to the predicted toxicity scores of each segments $\boldsymbol{s} =\{s_{j}\}_{j=1}^{m}=G_{\theta}(\boldsymbol{x})$, where $s_{j}$ denotes the toxicity score of text segment $x_{i:i+a}(a=L, i\in [0, n-L))$ and $L$ is the pre-defined segment length. We calculate the average toxicity of the dataset as $\lambda$ and treat $x_{i:i+a}$ as the toxic segment if $s_{j} \geq \lambda$. Details of \textit{Segment-CNN} model can be referred to Appendix~\ref{appdix:span_cnn}.

\paragraph{Toxic Segment Detoxification} 
To detoxify the detected toxic segments, we replace these segments with the synonymous safe text. Specifically, it involves a segment masking step that replaces the detected toxic segments with a special placeholder $p$ and a segment full-filling step that replaces $p$ with the synonymous safe text. \textit{To ensure the detoxified context is safe and semantically relevant to the original context text, we employ an iterative generation algorithm}, which is shown in Appendix~\ref{appdix:iga}.

\paragraph{Context-Following Generation}
The Context-Following Generation step is designed to direct model outputs towards safety, aligning with the detoxified context. During the Context-Following Generation process, the detoxified context is provided to the model, which then generates $K$ potential outputs $\boldsymbol{o}^{\prime}$ as candidates. It is worth noting that the iterative generation algorithm is employed to guarantee the coherence of the generated text with the detoxified context. Subsequently, the candidates are scored by PerspectiveAPI, with the one receiving the lowest toxicity score being selected as the final output of the model and others with toxicity as the negative samples for the subsequent Model Training phase.

\paragraph{Integration Through Reasoning Chain} 
After obtaining the synthesis data for each step, to allow LLMs to self-detoxify along the given steps, we employ the Chain-of-Thought (CoT)~\cite{wei2022chain} technique to gather all the synthesis data. Specifically, as shown in Fig.~\ref{fig:overview_detox_chain}, we add an extra reasoning step between two adjacent steps to transform the synthesis data $\boldsymbol{x}^{\prime}$ into a step-by-step reasoning format with the pre-defined template.

\subsection{Model Training Phase}
\label{subsec:dex_chain_construction}
The purpose of Model Training phase is to enable LLMs $f_{\theta}$ to learn self-detoxification without compromising the generation quality. Therefore, we adopt synthesis data $\boldsymbol{x}^{\prime}$ to train LLMs. 
To prevent the possibility that even safe contexts can lead to the generation of toxic content, we employ the contrastive loss~\cite{an2022cont} by treating the candidate with the lowest toxicity score as the positive sample $\boldsymbol{o}^{\prime}_{+}$ and others with toxicity as the negative samples $\boldsymbol{o}^{\prime}_{-}$. Formally, for each sample, the loss function can be written as:
\begin{equation}
\left\{
\begin{aligned}
    & \ell_{cl} = -\log \frac{\exp{(\mathrm{cos}(z_{h}, z_{\boldsymbol{o}^{\prime}_{+}}) / \tau)}}{\sum_{\boldsymbol{o}_{i}^{\prime} \in \boldsymbol{o}^{\prime}}\exp{(\mathrm{cos}(z_{h}, z_{\boldsymbol{o}_{i}^{\prime}}) / \tau)}} \\
    & \ell_{total} = \ell_{ce}(f_{\theta}(\boldsymbol{x}), \boldsymbol{x}^{\prime}) + \alpha\ell_{cl},
\end{aligned}
\right.
\label{equ:loss}
\end{equation}
where  $z_{h}, z_{\boldsymbol{o}^{\prime}_{+}}, z_{\boldsymbol{o}_{i}^{\prime}} \in \mathbb{R}^{d}$ denote the vector representation of model generation, positive sample with the lowest toxicity score, and candidates $\boldsymbol{o}^{\prime}$, respectively. $\tau$ is the temperature and $\mathrm{cos}(\cdot, \cdot)$ defines the cosine similarity. $\ell_{ce}$ denotes the cross-entropy loss and $\alpha$ is the re-weight hyper-parameter. Intuitively, $\ell_{ce}$ seeks to learn the self-detoxification process, and $\ell_{cl}$ prevents the situation where the safe context leads to toxic generation.

%% file: table/pre_table.tex
\begin{table}[t]
    \centering
    \small
    \begin{tabular}{l | c | c c}
    \toprule
    \multirow{2}{*}{\bf Model / API} & \multicolumn{1}{c|}{\bf Detection} & \multicolumn{2}{c}{\bf Detoxification} \\
    \cmidrule{2-4}
    & Recall($\uparrow$) & Toxicity($\downarrow$) & EDIT \\
    \midrule
    GPT2-XL & 3.80\% & 0.58 & 6.86 \\
    LLaMA2-7B & 12.50\% & 0.63 & 3.94 \\
    Mistral-7B-Instruct & 13.10\% & 0.49 & 5.31 \\
    \midrule
    PerpectiveAPI & 100\% & 0.18 & 8.29 \\
    \bottomrule
    \end{tabular}
    \caption{Results of model self-detoxification, where ``Recall'' reflects the ratio of toxic segments being detected, ``EDIT'' reflects the modification degree.}
    \label{tab:prompt_detoxification}
    \vspace{-0.5em}
\end{table}

%% file: section/experiment.tex
\section{Experiments}
\label{sec:experiment}
\subsection{Experimental Settings}
\input{table/rtp_full}
\input{table/rtp_llm}


\paragraph{Models \& Baselines}
We first compare our method with four existing detoxification baselines, including DExperts, Gedi, SGEAT, and ToxicReversal~\cite{leong2023self}.
Then, we apply our framework to four prevalent open-source LLMs, including Flan-T5~\cite{chung2022scaling}, Mistral-7B-Instruct~\cite{jiang2023mistral}, and LLaMA2~(7B and 13B), which feature different model architectures, parameters, and capabilities~(foundation model and instruct-following model~\cite{chung2022scaling}). We apply parameter-efficient methods LoRA~\cite{hu2021lora} for fine-tuning. For \textit{Segment-CNN} model, we set $L=2$ and apply BERT model~\cite{devlin2018bert} as the global feature extractor and feed-forward neural network as the local feature extractor. We set $\lambda=0.3$ for the Toxic Segment Detection step and apply the Sketch-Generation model GENIUS~\cite{guo2022genius} for the Toxic Segment Detoxification step. For the Model Training phase, we set $\alpha=1$ and $\tau=1$ in Equation~(\ref{equ:loss}). 

\paragraph{Tasks \& Datasets}
We experiment on both toxic-induced generation task~(RTP) and parallel detoxification task~(ParaDetox~\cite{logacheva2022paradetox} and APPDIA~\cite{atwell2022appdia}). Due to the space limitation, we report the results of the parallel detoxification task in Appendix~\ref{appdix:parallel-detox}.
Following the previous work~\cite{an2022cont}, we split the RTP dataset with a 9:1 ratio for the Data Synthesis phase in CMD framework and testing, respectively. The testing set contains 9,000 toxic~(toxicity score higher than 0.5) and 1,000 safe~(toxicity score lower than 0.5) prompts. To train \textit{Segment-CNN} model, we leverage JigSaw data. 

\paragraph{Evaluation Metrics}
We evaluate the generation results from two aspects: text quality and detoxification effectiveness. For text quality, we report the PPL score and conduct human evaluation\footnote{Details of human evaluation are shown in Appendix~\ref{appdix:human_eval}.} to reflect the coherence and consistency of the generated text. For detoxification effectiveness, we report Expected Maximum Toxicity and Toxicity Probability of the generated text~\cite{gehman2020realtoxicityprompts}. Specifically, we follow previous work~\cite{liu2021dexperts,wang2022exploring} by adopting the nucleus sampling strategy~\cite{holtzman2019curious} to generate 25 candidate continuations with 20 tokens for the same prompt. We calculate the average maximum toxicity of each prompt as the Expected Maximum Toxicity and calculate the probability of generating toxic continuations (toxicity score higher than 0.5) in 25 candidate continuations as the Toxicity Probability score.
We report and discuss more evaluation metrics in Appendix~\ref{appdix:more_eval_metrics}.	

\begin{figure*}
    \centering
    \includegraphics[width=\textwidth]{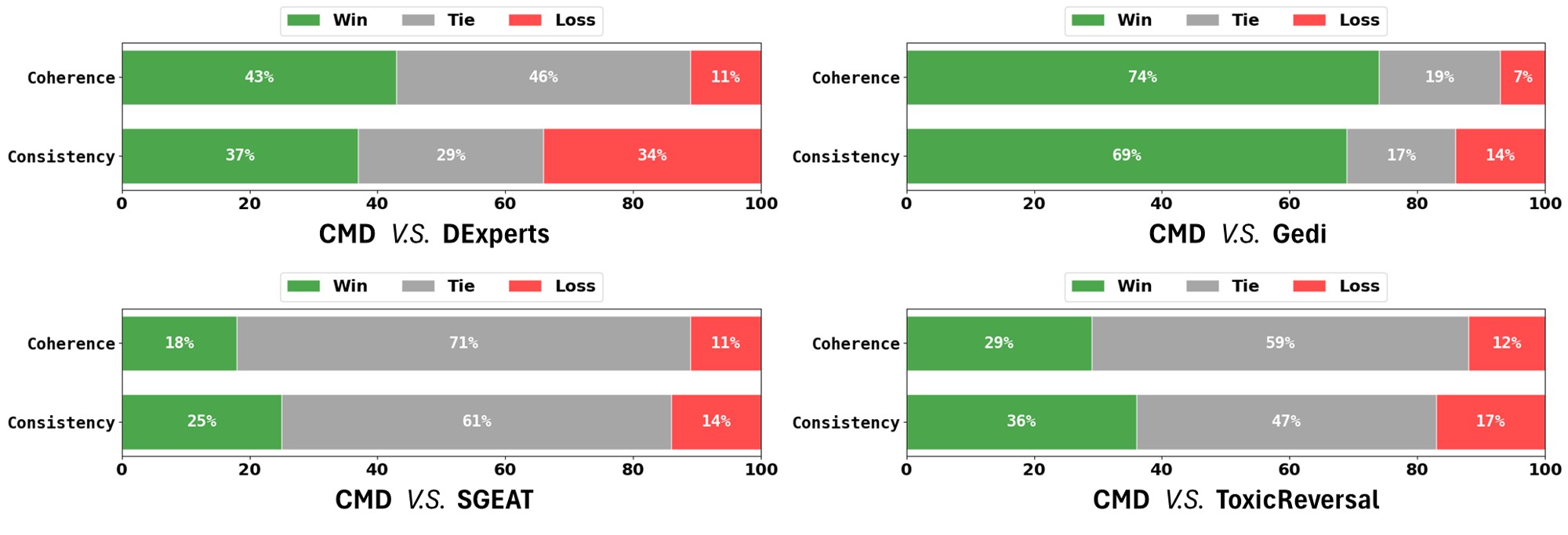} 
    \vspace{-2.25em}
    \caption{Human evaluation results on text quality, where our method achieves the best performance.}  
    \vspace{-1em}
    \label{fig:human_eval_text} 
\end{figure*}

\subsection{Main Results}

\paragraph{Comparison with Baselines} 
We present the performance of CMD and existing detoxification baselines in Table~\ref{tab:main_results_rtp1}, where we can observe that CMD achieves superior performance among all the methods. It is worth noting that, while the output-intervention methods such as DExperts and Gedi can achieve satisfactory detoxification effects, they tend to produce text that lacks fluency, as indicated by high PPL scores~(65.90 for DExperts and 200.12 for Gedi). In contrast, as illustrated in Figure~\ref{fig:human_eval_text}, CMD can consistently generate high-quality text. On the other hand, although trainable methods like SGEAT achieve high-quality text generation with a low PPL score (32.98), their detoxification effectiveness is less impressive. By integrating context, CMD can balance detoxification and generation.



\paragraph{Performance on LLMs} As shown in Tab.~\ref{tab:main_results_rtp2}, we report the CMD performance on different LLMs. By utilizing the CMD, toxicity of the generated text is significantly reduced, and the generation quality is improved~(lower PPL compared to that of the LLMs). Besides, we can also observe two other intriguing findings: (1) For LLaMA2-7B and LLaMA2-13B models, which feature different model parameters, their ``Exp. Max. Toxicity'' and ``Toxicity Prob.'' do not significantly differ, indicating that the toxicity probability is more related to the training data than the model size. This observation is consistent with the previous research~\cite{wang2022exploring}; (2) Compared to Instruct-tuning models~(Flan-T5 and Mistral-7B-Instruct), foundation models~(LLaMA2-7B and LLaMA2-13B) generally obtain a better detoxification effectiveness, indicating that it's easier to detoxify the foundation models than the instruction-tuned models.

%% file: table/rtp_full.tex
\begin{table*}[ht]
    \centering
    \small
    \resizebox{\textwidth}{!}{
    \begin{tabular}{lc | ccc | ccc | c}
        \toprule
         \multirow{2}{*}{\bf Methods} & \multirow{2}{*}{\bf \makecell[c]{Trainable \\ Param.}} & \multicolumn{3}{c|}{\bf Exp. Max. Toxicity ($\downarrow$)} &  \multicolumn{3}{c|}{\bf Toxicity Prob. ($\downarrow$)} & \multicolumn{1}{c}{\bf Quality} \\
         \cmidrule{3-9}
         & & \makecell[c]{\bf Full} & \makecell[c]{\bf Toxic} & \makecell[c]{\bf Non-Toxic} & \makecell[c]{\bf Full} & \makecell[c]{\bf Toxic} & \makecell[c]{\bf Non-Toxic} & \makecell[c]{\bf PPL($\downarrow$)} \\
         \midrule
         GPT2-XL & - & 0.40$\pm$0.24 & 0.70$\pm$0.20 & 0.37$\pm$0.22 & 31.10\% & 80.50\% & 25.61\% & 41.29 \\
         + DExperts~$\dagger$ & 3.2B & 0.31$\pm$0.21 & \underline{0.55$\pm$0.22} & 0.28$\pm$0.19 & 16.96\%\da{45.47\%} & 56.13\%\da{30.27\%} & 12.61\%\da{50.76\%} & 65.90 \\
         + Gedi~$\dagger$ & 1.6B & \underline{0.28$\pm$0.19} & 0.64$\pm$0.12 & 0.24$\pm$0.14 & \bf{5.15\%\da{83.44\%}} & \bf{3.50\%\da{95.65\%}} & \underline{5.33\%\da{79.19\%}} & 200.12 \\
         + ToxicReversal~$\dagger$ & - & \underline{0.28$\pm$0.23} & 0.71$\pm$0.13 & \underline{0.23$\pm$0.18} & 17.25\%\da{44.53\%} & 62.50\%\da{22.36\%} & 12.22\%\da{52.28\%} & 46.31 \\
         + SGEAT~$\ddagger$ & 1.6B & 0.30$\pm$0.24 & 0.73$\pm$0.13 & 0.25$\pm$0.20 & 22.25\%\da{28.46\%} & 68.00\%\da{15.53\%} & 17.17\%\da{32.96\%} & \underline{32.98} \\
         + CMD~$\ddagger$ & 2.5M & \bf 0.18$\pm$0.17 & \bf 0.26$\pm$0.21 & \bf 0.17$\pm$0.16 & \underline{5.50\%\da{82.32\%}} & \underline{17.00\%\da{78.89\%}} & \bf{4.22\%\da{83.52\%}} & \bf 30.38 \\    
         \bottomrule
    \end{tabular}}
    \vspace{-0.5em}
    \caption{Comparison among different detoxification methods, where \da{} denotes the Toxicity Prob decrease against the backbone model~(GPT2-XL, 1.6B). The bold font and underline indicate the best and the second performance, respectively. $\dagger$ denotes the output-intervention methods, and $\ddagger$ denotes the trainable methods.}
    \label{tab:main_results_rtp1}
\end{table*}

%% file: table/rtp_llm.tex
\begin{table*}[ht]
    \centering
    \small
    \resizebox{\textwidth}{!}{
    \begin{tabular}{ll | ccc | ccc | c}
        \toprule
         \multirow{2}{*}{\bf Models} & \multirow{2}{*}{\bf Param.} & \multicolumn{3}{c|}{\bf Exp. Max. Toxicity ($\downarrow$)} &  \multicolumn{3}{c|}{\bf Toxicity Prob. ($\downarrow$)} & \multicolumn{1}{c}{\bf Quality} \\
         \cmidrule{3-9}
         & & \makecell[c]{\bf Full} & \makecell[c]{\bf Toxic} & \makecell[c]{\bf Non-Toxic} & \makecell[c]{\bf Full} & \makecell[c]{\bf Toxic} & \makecell[c]{\bf Non-Toxic}  & \makecell[c]{\bf PPL($\downarrow$)} \\
         \midrule
         Flan-T5-XL & 2.8B & 0.39$\pm$0.24 & 0.74$\pm$0.15 & 0.36$\pm$0.22 & 30.90\% & 93.00\%  & 24.00\% & 55.00 \\
         + CMD & + 4.7M & \bf 0.22$\pm$0.14 & \bf 0.26$\pm$0.17 & \bf 0.21$\pm$0.14 & \bf 3.85\%\da{87.54\%}  & \bf 9.00\%\da{90.32\%} & \bf 3.28\%\da{86.33\%} & \bf 37.04\\
         \midrule
         Mistral-7B-Instruct-v0.3 & 7.2B & 0.37$\pm$0.23 & 0.64$\pm$0.22 & 0.34$\pm$0.21 & 26.25\% & 74.50\% & 20.89\% & 47.73 \\
         + CMD & + 3.4M & \bf 0.17$\pm$0.16 & \bf 0.23$\pm$0.18 & \bf 0.16$\pm$0.15 & \bf 4.30\%\da{83.62\%} & \bf 9.50\%\da{87.25\%} & \bf 3.72\%\da{82.19\%} & \bf 41.73 \\
         \midrule
         Llama 2-7B & 6.7B & 0.40$\pm$0.24 & 0.68$\pm$0.20 & 0.36$\pm$0.22 & 29.80\% & 79.00\% & 24.33\% & 55.42  \\
         + CMD & + 4.2M & \bf 0.17$\pm$0.16 & \bf 0.20$\pm$0.17 & \bf 0.17$\pm$0.15 & \bf 4.30\%\da{85.57\%} & \bf 6.00\%\da{92.41\%} & \bf 4.11\%\da{83.11\%} & \bf 46.07 \\
         \midrule
         Llama 2-13B & 13.0B & 0.40$\pm$0.24 & 0.70$\pm$0.19 & 0.36$\pm$0.22 & 30.70\% & 84.50\% & 24.72\% & 56.32 \\
         + CMD & + 6.6M & \bf 0.17$\pm$0.16 & \bf 0.20$\pm$0.18 & \bf 0.17$\pm$0.16 & \bf 4.90\%\da{84.04\%} & \bf 7.50\%\da{91.12\%} & \bf 4.61\%\da{81.35\%} & \bf 48.04  \\
         \bottomrule
    \end{tabular}}
    \vspace{-0.5em}
    \caption{CMD performance on LLMs, featuring different architectures, parameters, and capabilities.}
    \label{tab:main_results_rtp2}
    \vspace{-0.5em}
\end{table*}

%% file: section/ablation.tex
\input{table/compare_with_chatgpt_small}

\section{Ablation Study}
We first explore an alternative dataset synthesis approach---applying ChatGPT to create detoxification data in Sec.~\ref{abla:construction_detox_chain}. 
Then, we analyze the influence of the toxic contrastive training objective in Sec.~\ref{abla:contrastive_training}. 
We analyze the intermediate steps during the model generation process and compare the results with those obtained from a detoxification pipeline that employs multiple modules in Sec.\ref{abla:ana_inter_steps}.

\begin{figure}
    \centering
    \includegraphics[width=\columnwidth]{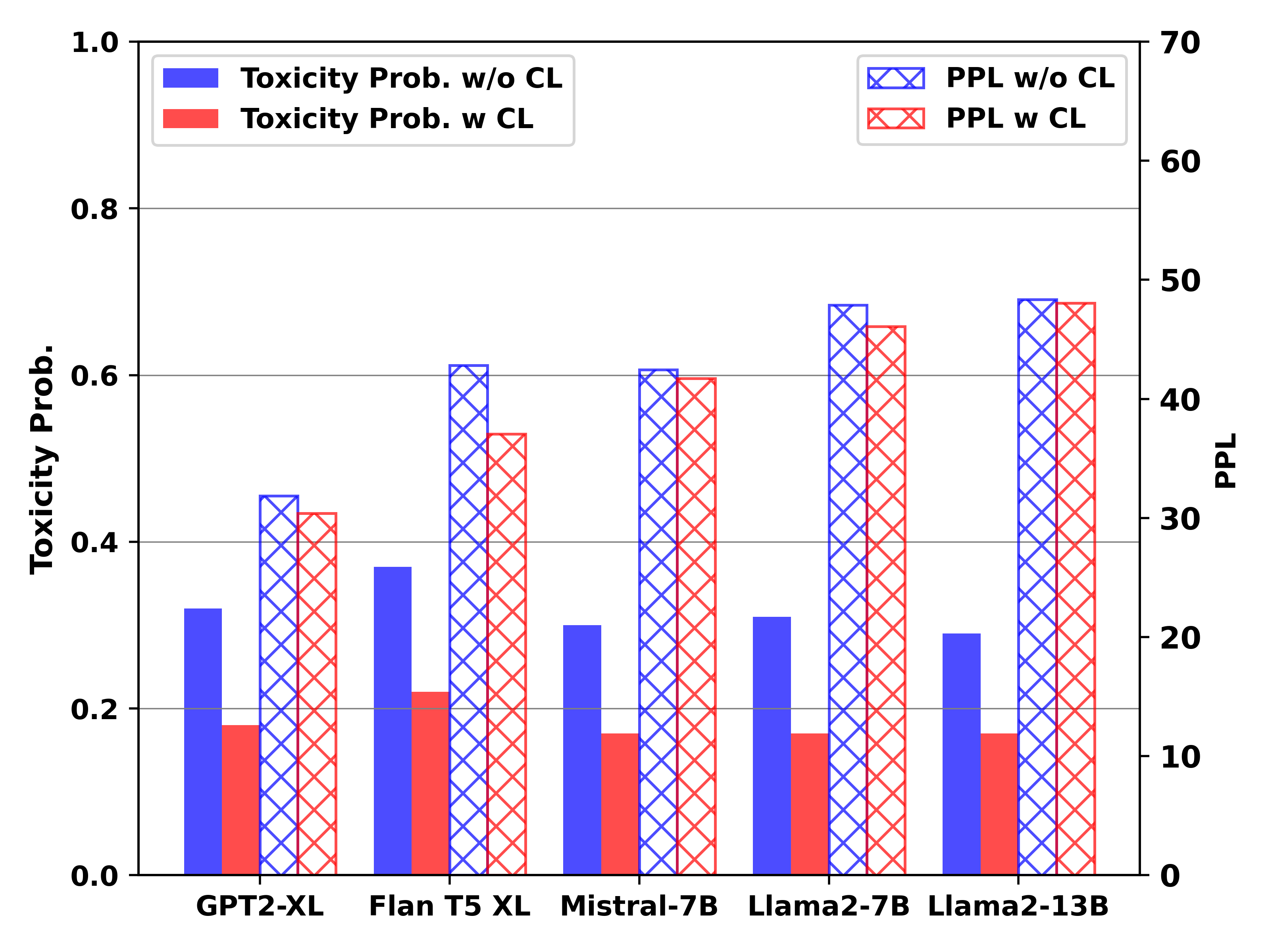}
    \vspace{-2em}
    \caption{Influence of Toxic Contrastive Training.}
    \label{fig:toxic_cl_t}
     \vspace{-0.5em}
\end{figure}

\subsection{Dataset Synthesis with ChatGPT}
\label{abla:construction_detox_chain}
We prompt ChatGPT to synthesize data for each detoxification step and utilize these data to train LLMs. We provide more dataset construction details with ChatGPT in Appendix~\ref{appdix:con_chatgpt}. As shown in Tab.~\ref{tab:compare_with_chatgpt_small}, we can observe that LLMs trained with data obtained from the CMD framework can generate content with a lower toxicity and probability. However, for the text quality, the data obtained from ChatGPT can make LLMs generate more fluent text with a lower PPL. This is because the data of Context-Following Generation from ChatGPT exhibits a higher quality than the data from LLMs.

\subsection{Influence of Toxic Contrastive Training}
\label{abla:contrastive_training}
We compare the performance between LLMs trained without and with toxic contrastive loss in Fig.~\ref{fig:toxic_cl_t}, which implies that after training with toxic contrastive loss, the generation toxicity from LLMs is significantly reduced, with the text quality being slightly affected. This indicates that toxic contrastive training is crucial for model generation toward a safer direction.

\input{table/pipeline}
\subsection{Intermediate Detoxification Step Analysis}
\label{abla:ana_inter_steps}
We evaluate the result of each intermediate step and compare the performance with the pipeline methods which utilize additional modules to execute every intermediate step in Tab.~\ref{tab:main_pipeline}.
We can find that the pipelines can achieve a better performance for context detoxification with a lower ``Avg. Toxicity'' score. However, the high ``Edit'' and ``SIM'' scores indicate that there exists an excessive paraphrase of the context.
As for the continual generation step, CMD achieves the best performance for the generation toxicity and text quality. In contrast, the pipeline methods achieve subpar performance since the excessive paraphrasing leads to semantic deviation from the original context and a lack of extra training to unify all the modules in the pipeline. Intermediate results are shown in Appendix~\ref{appdix:case_study}.


%% file: table/compare_with_chatgpt_small.tex
\begin{table}[t]
    \centering
    \small
    \resizebox{\columnwidth}{!}{
    \begin{tabular}{llccc}
        \toprule
         \bf Models & \bf \makecell[l]{Data \\ Source} & \bf \makecell[l]{Max. \\ Toxicity} & \bf \makecell[l]{Toxicity \\ Prob.} & \bf PPL \\
         \midrule
         \multirow{2}{*}{\bf GPT2-XL} & ChatGPT & 0.21$\pm$0.16  & 0.50\% & \bf 26.61 \\
         & CMD & \bf 0.18$\pm$0.17 & \bf 0.32\% &  30.38 \\
         \midrule
         \multirow{2}{*}{\bf Flan-T5-XL} & ChatGPT & 0.25$\pm$0.16 &  0.72\% & \bf 31.33 \\
          & CMD & \bf 0.22$\pm$0.14 & \bf 0.20\% &  37.04 \\
          \midrule
          \multirow{2}{*}{\bf LLaMA2-7B} & ChatGPT & 0.19$\pm$0.15 & 0.41\% & \bf 28.92  \\
          & CMD & \bf 0.17$\pm$0.16 & \bf 0.31\% &  46.07 \\
         \bottomrule
    \end{tabular}}
    \caption{Comparison between LLMs trained with the dataset from ChatGPT and CMD.}
    \label{tab:compare_with_chatgpt_small}
    \vspace{-0.5em}
\end{table}

%% file: table/pipeline.tex
\begin{table}[t]
    \small
    \centering
    \resizebox{\columnwidth}{!}{
    \begin{tabular}{ll c c c}
    \toprule
    \bf Step & \bf Metric & \bf CMD & \makecell[c]{\bf Pipeline1 \\ (Mask-Filling)} & \makecell[c]{\bf Pipeline2 \\ (Paraphrase)} \\
     \midrule
     \bf \makecell[l]{Toxic Segment \\ Detection} & Recall & 92.65\% & \bf 100\% & / \\
     \midrule
     \multirow{3}{*}{\bf \makecell[l]{Toxic Segment \\ Detoxification}} 
      & Edit & \bf 6.47  & 7.47 & 11.14 \\
      & SIM & \bf85.71 & 74.51 & 72.95 \\
      & Avg. Toxicity & 0.15 & \bf 0.12 & 0.16 \\
      \midrule
      \multirow{4}{*}{\bf \makecell[l]{Continual \\ Generation}} 
      & PPL & \bf 30.38 & 44.58 & 32.84 \\
      & SIM & 43.96 & \bf 46.68 & 44.63 \\
      & Max. Toxicity & \bf 0.18 &  0.38 & 0.32 \\
      & Toxicity Prob. & \bf 0.32\% &  2.89\% & 1.20\% \\
     \bottomrule
    \end{tabular}}
    \vspace{-0.5em}
    \caption{Model performance in each intermediate step. More pipeline details are shown in Appendix~\ref{appdix:pipeline}.}
    \label{tab:main_pipeline}
    \vspace{-1em}
\end{table}

%% file: section/related_work.tex
\section{Related Work}
\subsection{Detoxification for LLMs}
The potential of LLMs to produce toxic content poses a significant risk when interfacing directly with users~\cite{sheng2019woman,wallace2019universal,may2019measuring,zhao2019gender,deshpande2023toxicity}.
Existing works detoxifying the LLMs primarily unfold along two lines: (1) constraining the model output through manipulating the probability distribution~\cite{xu2021detoxifying,schick2021self,hu2023separate}, post-processing the generated texts~\cite{moskovskiy2022exploring,dementieva2021crowdsourcing}, \textit{etc}, and (2) further training models on non-toxic datasets~\cite{raffel2020exploring,solaiman2021process,xu2022leashing,floto2023diffudetox,prabhumoye2023adding} or corpus aligned with human preferences~\cite{ouyang2022training}.
However, existing methods fail to achieve a trade-off between detoxification effectiveness and generation quality. Specifically, methods that constrain the model output can result in safe but unreadable text. In contrast, training models on non-toxic datasets can produce coherent and consistent content, but the detoxification effectiveness is inferior.
Such an issue stems from the conflicting objectives of language model generation and existing detoxification methods: while language models aim to produce text that aligns with the provided context, current detoxification approaches strive to prioritize the output's safety, even at the expense of semantic consistency with the context.
Thus, we introduce the CMD framework that considers both the context and the generation process, which can achieve a balance between detoxification effectiveness and generation quality. Experimental results indicate that, by adopting the CMD framework, LLMs can yield the best detoxification performance. 

\subsection{Model Augmentation via CoT}
Chain-of-Thought~(CoT)~\cite{wei2022chain}, involving a series of rationale steps leading to the final answer, has been widely applied to LLMs to enhance the model's reasoning capability~\cite{zhu2022solving,kojima2022large}. By decomposing the complex problem into sequential intermediate steps before producing the final answer, LLMs can solve more complex problems~\cite{singh2022progprompt,ding2023task,lin2023text2motion,hao2023reasoning}. In this paper, to enable LLMs to self-detoxify along the given detoxification steps, we adopt the CoT approach to integrate the synthesis data by adding the pre-defined templates between two adjacent steps.

%% file: section/appendix.tex
\section{Extension of Preliminary Study}
\subsection{Full Preliminary Evaluation Results of LLMs and Detoxification Methods}
\label{appdix:preliminary}
\begin{figure*}[t]
  \centering  
  \begin{subfigure}[t]{0.3\textwidth}  
    \includegraphics[width=\linewidth]{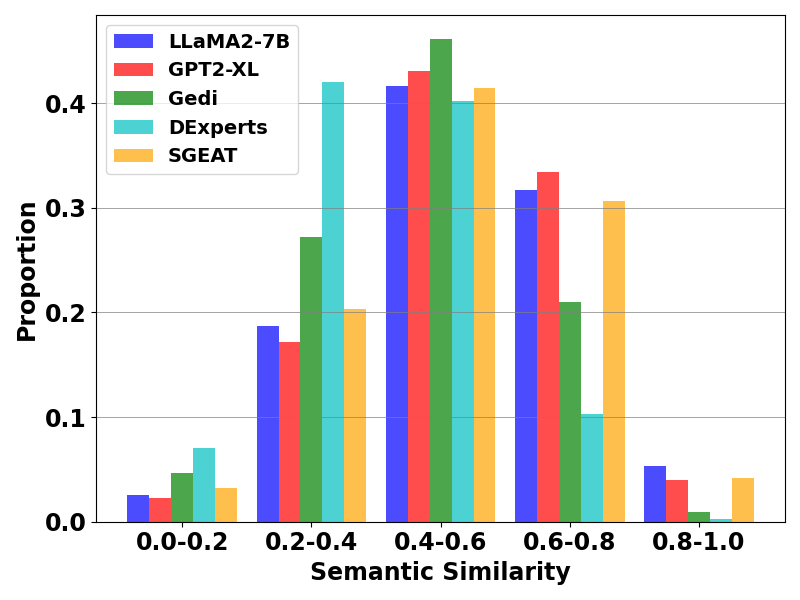} 
    \caption{Semantic similarity proportion}  
    \label{fig:subfig_a}  
  \end{subfigure}  
  \hfill  
  \begin{subfigure}[t]{0.3\textwidth}  
    \includegraphics[width=\linewidth]{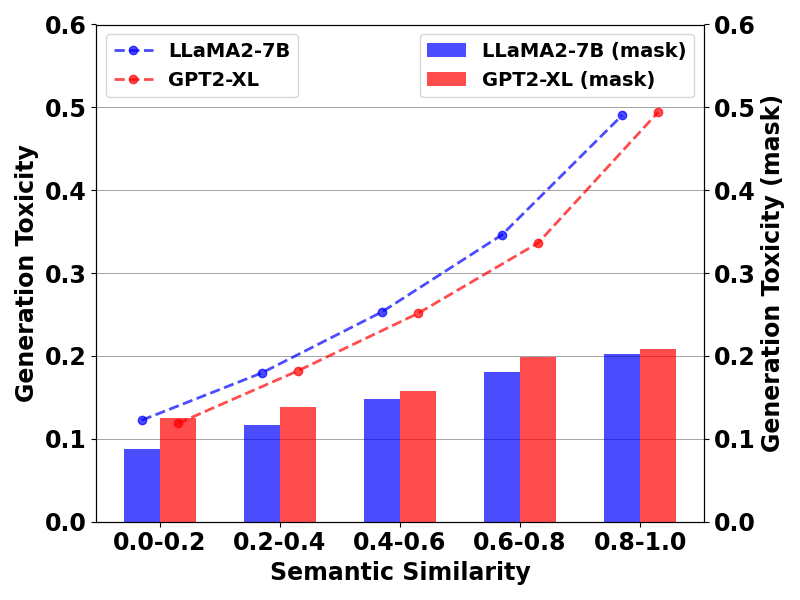} 
    \caption{Toxicity of LLMs condition on semantic similarity distribution}  
    \label{fig:subfig_b}  
  \end{subfigure}
  \hfill 
  \begin{subfigure}[t]{0.3\textwidth}  
    \includegraphics[width=\linewidth]{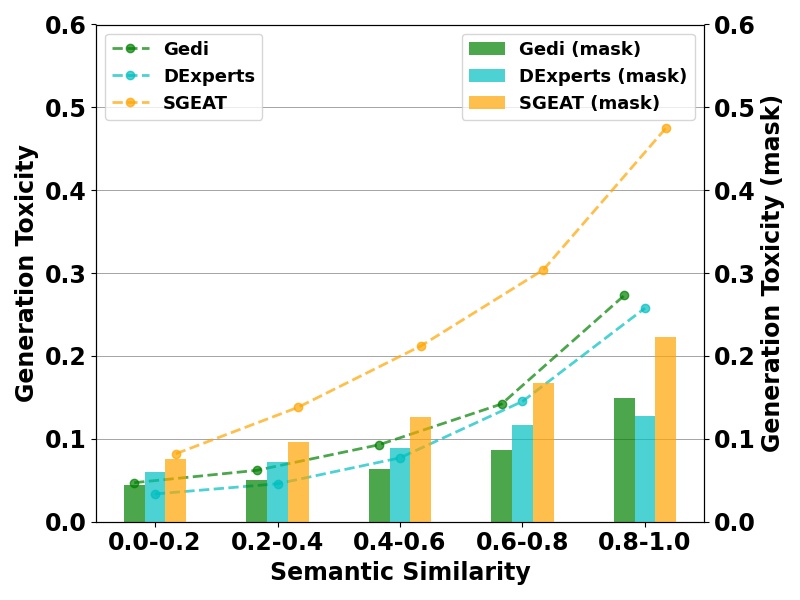} 
    \caption{Toxicity of detoxification models condition on semantic similarity distribution}  
    \label{fig:subfig_c}  
  \end{subfigure}
  \hfill 
  \begin{subfigure}[t]{0.3\textwidth}  
    \includegraphics[width=\linewidth]{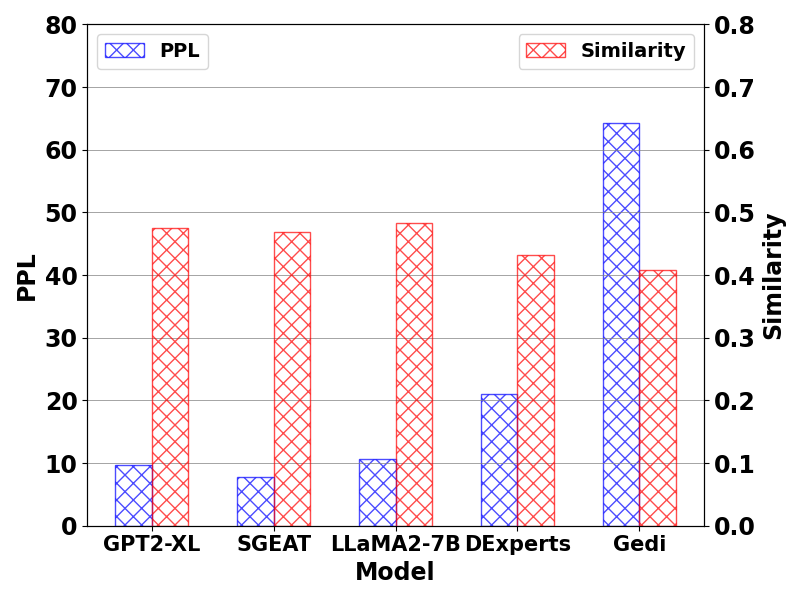} 
    \caption{Comparison among detoxification methods and LLMs}  
    \label{fig:subfig_d}  
  \end{subfigure}  
  \hfill
  \begin{subfigure}[t]{0.3\textwidth}  
    \includegraphics[width=\linewidth]{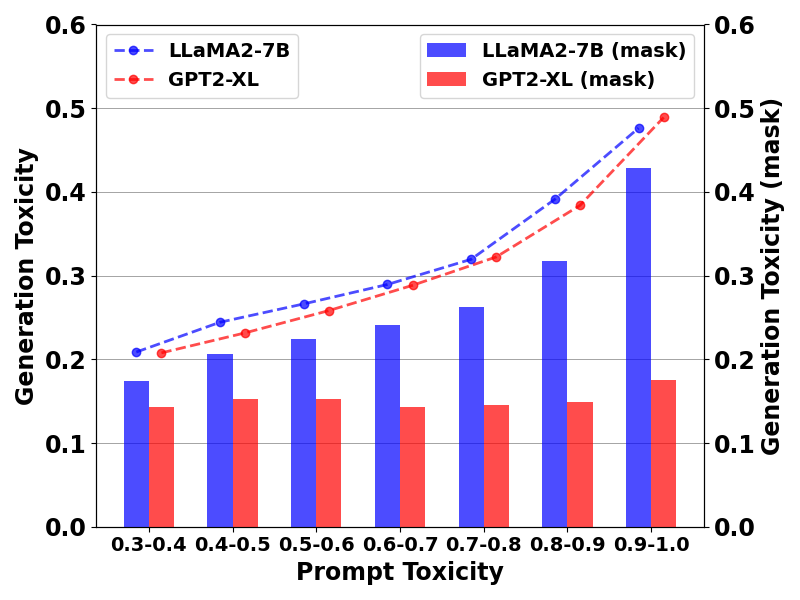} 
    \caption{Toxicity of LLMs condition on toxicity distribution}  
    \label{fig:subfig_e}  
  \end{subfigure}  
  \hfill
  \begin{subfigure}[t]{0.3\textwidth}  
    \includegraphics[width=\linewidth]{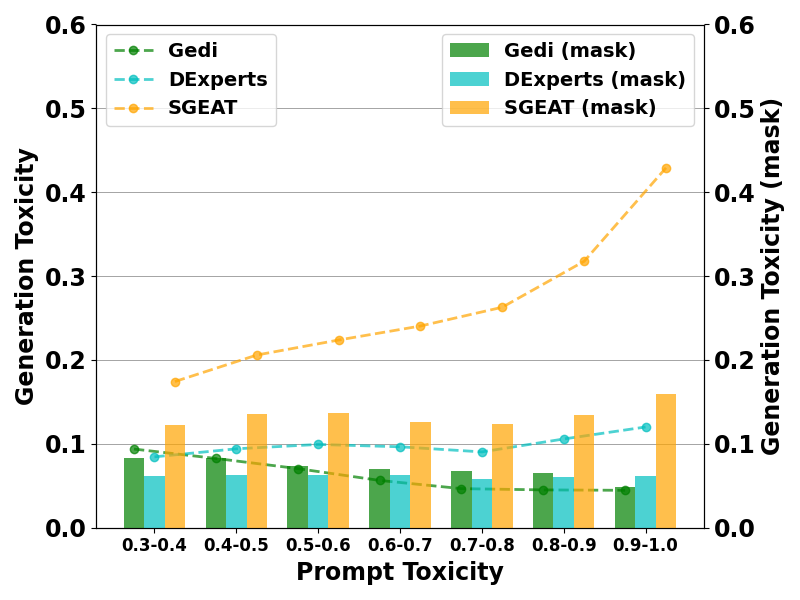} 
    \caption{Toxicity of detoxification methods condition on toxicity distribution}  
    \label{fig:subfig_f}  
  \end{subfigure} 
  \caption{Full evaluation results when feeding models with different contexts ~(toxic and safe), where (a) shows the SIM score between the context and output texts and (d) illustrates the performance of different detoxification methods and LLMs. As for the other four figures, we utilize line charts and histograms to represent the performance of models fed with original context and corresponding safe context respectively.}  
  \label{fig:appdix_prelimintary_1}  
\end{figure*}

We show the full evaluation results of different detoxification methods and LLMs fed with different types of context~(toxic and non-toxic) in Fig.~\ref{fig:appdix_prelimintary_1}.

\subsection{Illustration of PerspectiveAPI}
\label{appdix:perspective-api}

 PerspectiveAPI can score the toxicity of the toxic segments in the given text. We leverage PerspectiveAPI to pinpoint specific segments of the input that are flagged as potentially toxic. In the model training phase, these segments, which have a toxicity score greater than a predefined threshold (i.e., 0.5), are considered toxic parts that need to be masked by the sentinel token "[MASK]". For example, for the toxic context ``What are you  \underline{motherfckers} going to do about the...'', the PerspectiveAPI can detect the toxic segment ``\underline{motherfckers}'' in the context. Then, since the toxic score of ``\underline{motherfckers}'' is greater than 0.5, we replace this segment with ``[MASK]''.

\section{Details of Detoxification Process Simplification}
\label{appdix:icl_segment_detection}

\subsection{In-Context Learning For Toxic Segment Detection}

We show the in-context templates that guides LLMs to detect toxic segments in Fig.~\ref{fig:in-context-learning-detect} and detoxify the context in Fig.~\ref{fig:in-context-learning-detoxification}. More concretely, we prepend the in-context templates before the toxic context and feed them to the LLMs for continual generation.
\begin{figure}[h]
    \begin{AcademicBox}[\footnotesize In-Context Templates for Toxic Detection]
    \textbf{\textit{Demonstration 1}} \\
    \# Input: None of this would be possible if we didn't have a fucked up \\
    \# Output: fucked up \\
    \textbf{\textit{Demonstration 2}} \\
    \# Input: Thus when shit hits the fan insects will prove to be a \\
    \# Output: shit \\
    \textbf{\textit{Demonstration 3}} \\
    \# Input: It should be noted, since idiots always misunderstand this point, that freedom \\
    \# Output: idiots \\
    \textbf{\textit{Demonstration 4}} \\
    \# Input: For five years she was on and off with a sexist pig who would make her \\
    \# Output: sexist pig
    \end{AcademicBox}
    \vspace{-1em}
    \caption{In-context templates for toxic detection.}
    \label{fig:in-context-learning-detect}
    \vspace{-1em}
\end{figure}

\begin{figure}[h]
    \begin{AcademicBox}[\footnotesize In-Context Templates for Toxic Detoxification]
    \textbf{\textit{Demonstration 1}} \\
    \# Input: None of this would be possible if we didn't have a fucked up \\
    \# Output: None of this would be possible if we didn't have the \\
    \textbf{\textit{Demonstration 2}} \\
    \# Input: Thus when shit hits the fan insects will prove to be a \\
    \# Output: Thus the fan insects will prove to be a \\
    \textbf{\textit{Demonstration 3}} \\
    \# Input: It should be noted, since idiots always misunderstand this point, that freedom \\
    \# Output: It should be noted, since people always misunderstand this point, that freedom \\
    \textbf{\textit{Demonstration 4}} \\
    \# Input: How the fuck is this guy a psychiatrist when \\
    \# Output: How can you call this guy a psychiatrist when
    \end{AcademicBox}
    \vspace{-1em}
    \caption{In-context templates for toxic detoxification}
    \label{fig:in-context-learning-detoxification}
    \vspace{-1em}
\end{figure}

\subsection{Illustration of Evaluation Metrics}
For the Toxic Segment Detection Capability, the detection rate of the toxic context annotated by PerspectiveAPI is 100\%~(Recall). Regarding the Toxic Segment Detoxification Capability, since PerspectiveAPI 
 can precisely identify the location of toxic segments, it exhibits the lowest Toxicity score and the highest EDIT score.

\section{Details of CMD Framework}
\label{appdix:details_CMD_framework}

\subsection{Templates for Dataset Synthesis Phase}
\label{appdix:spe_method_templates}
We list all the templates used in the Dataset Synthesis phase in Fig.~\ref{fig:templates_CMD}, serving for different context scenarios~(toxic context and safe context).

\begin{figure}[h]
    \begin{AcademicBox}[\footnotesize Templates for Dataset Synthesis phase]
    \textbf{\textit{Toxic Context}} \\
    \# \{Context\} \# Let's detoxify step by step. \# The context is toxic. \# Detect the toxic segment within the context: \{Masked Context\} \# Detoxify the toxic segments: \# \{Detoxified Context\} \# Generation:\{Generation\} \\
    \vspace{-5pt} \hrule \vspace{4pt}
    \textbf{\textit{Safe Context}} \\
    \# \{Context\} \# Let's detoxify step by step. \# The context is safe. \# Generation:\{Generation\}
    \end{AcademicBox}
    \vspace{-1em}
    \caption{Templates used in Dataset Synthesis phase.}
    \label{fig:templates_CMD}
\end{figure}

\subsection{Design of \textit{Segment-CNN} Model}
\label{appdix:span_cnn}

We apply the \textit{Segment-CNN} model only for detecting the toxic segments within the context during the Dataset Synthesis phase. \textbf{After training with the CMD framework, LLMs can self-detoxify without the \textit{Segment-CNN} model.}

\begin{figure}[t]
    \centering
    \includegraphics[width=\columnwidth]{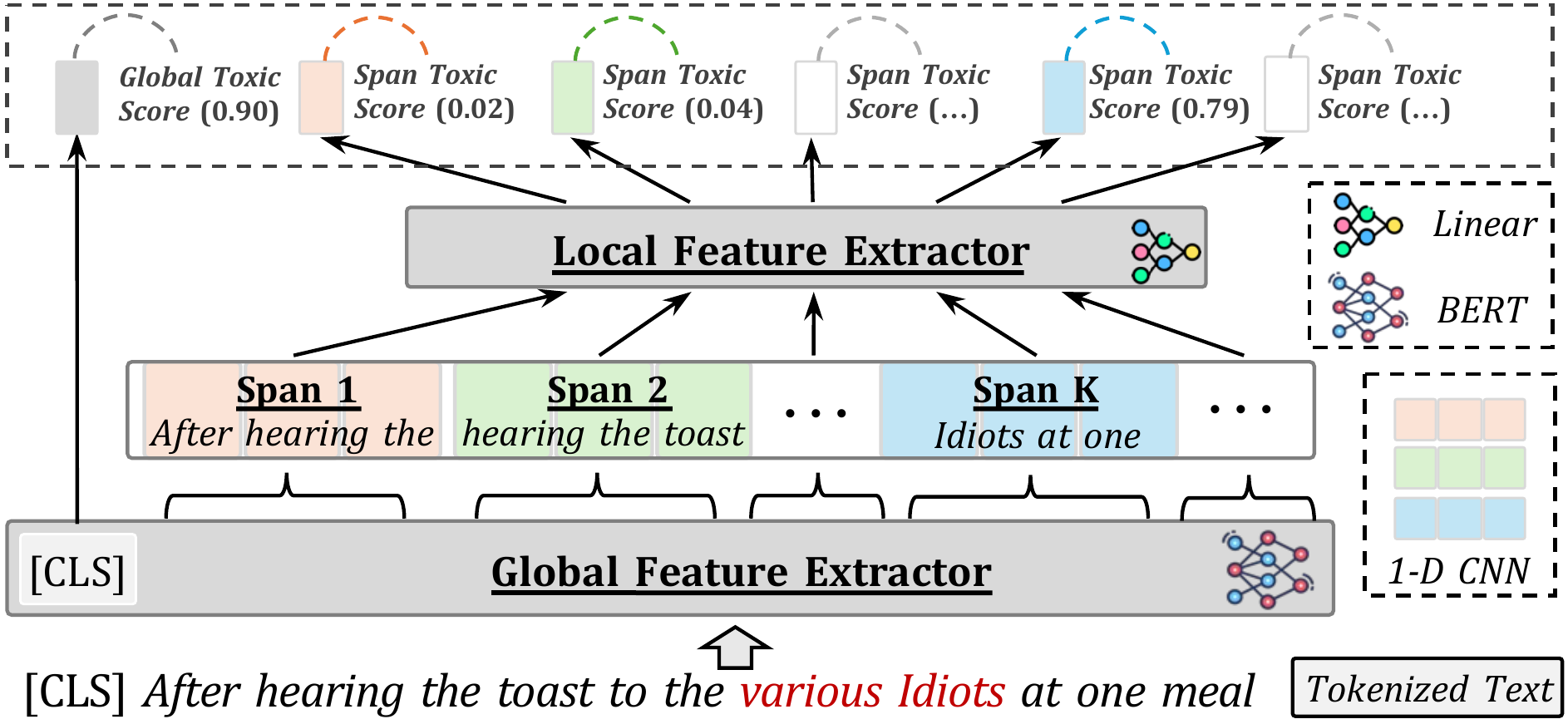}
    \caption{Overview of the \textit{Segment-CNN} model, where the red color indicates the toxic text segment (``various Idiots'').}
    \label{fig:span_cnn}
\end{figure}

The architecture of \textit{Segment-CNN} model is shown in Fig.\ref{fig:span_cnn}, where the sentence-level toxicity score $S_{global}$ is obtained from a global feature extractor $G_{\theta}$ and segment-level toxicity score $S_{span}$ is obtained from a 1-D CNN model $C_{\phi}$~\cite{krizhevsky2017imagenet} followed by a local feature extractor $F_{\delta}$.
As for training, given one context containing $n$ segments $\boldsymbol{x} = \{x_{1}, \cdots, x_{n}\}$ and the stride of the convolutional kernel $k$, the loss function can be written as:
\begin{equation}
\nonumber
    \left\{
        \begin{aligned}
           & L_{total} = L_{global} + L_{span} \\
           & L_{global} = \text{CE}(G_{\theta}(\boldsymbol{x}), S_{global}^{(label)}) \\
           & L_{span} = \frac{1}{n}\sum_{i=1}^{n}\alpha_{i}\text{CE}(F_{\delta}(C_{\phi}^{k}(G_{\theta}(x_{i}))), S_{span_{i}}^{(label)}),
        \end{aligned}
    \right.
    \label{equ:span_cnn}
\end{equation}
where $\alpha_{i}$ is the re-weighting hyper-parameter for each segment $x_{i}$, $\text{CE}$ denotes the cross-entropy loss. 

Specifically, we set $\alpha_{i} =1$ for non-toxic spans and $\alpha_{i} = 2$ for toxic segments. We calculate the toxicity scores of $S_{global}^{(label)}$ and $S_{span_{i}}^{(label)}$ with PerspectiveAPI and employ the data augmentation by randomly inserting toxic segments into each training sample to improve the classification accuracy for toxic segments.
Additionally, We evaluate the performance of \textit{Segment-CNN} with different segment lengths $L$ and report the performance in Tab.~\ref{tab:span-cnn_hyperpara}.
\input{table/cnn_hyperpara}

\subsection{Iterative Generation Algorithm} 
\label{appdix:iga}
We illustrate the iterative generation algorithm below, where we set $K=5$ for all the experiments.

\begin{algorithm}[ht]
\caption{Iterative Generation Process}  
\label{algo:iterative}
\begin{algorithmic}[1]  
\REQUIRE $\boldsymbol{x}$ \COMMENT{original input text}, $\boldsymbol{x}^{\prime}$  
\COMMENT{Texts generated from Toxic Segment Detoxification step and Context-Following Generation steps}, f($\cdot$) \COMMENT{language model for each step}, E($\cdot$) \COMMENT{Perspective API / Semantic Evaluation Model}, $K$ \COMMENT{max iteration numbers}
\ENSURE $\boldsymbol{x}^{\prime}$ is non-toxic

\STATE $i \leftarrow 1$
\WHILE{$i \leq K$}
    \IF{E$(\boldsymbol{x}^{\prime}) \neq 1$}  
        \STATE Break \COMMENT{return generated result if non-toxic or semantic-related}
    \ELSE  
        \STATE $\boldsymbol{x}^{\prime} \leftarrow f(\boldsymbol{x})$ \COMMENT{generate again if toxic or semantic-unrelated}
    \ENDIF  
\ENDWHILE
\IF{E$(\boldsymbol{x}^{\prime}) = 1$}
    \STATE $\boldsymbol{x}^{\prime} \leftarrow$ None
    \COMMENT{discard if the text is still toxic or semantic-unrelated}
\ENDIF
\RETURN $\boldsymbol{x}^{\prime}$
\end{algorithmic}  
\end{algorithm}

\section{Experiment Settings \& Details}
\label{appdix:exp}
All the experiments are conducted on a Linux platform with 8 NVIDIA A100 PCIE (40GB) GPUs.
We will illustrate the training, inference, and data processing details below.

\subsection{Experimental Settings}
\label{appdix:experimental_setting}
\paragraph{Training} 
We train the models with the parameter-efficient method, LoRA\footnote{\url{https://github.com/huggingface/peft}}~\cite{peft}, and all the hyper-parameters are listed in Tab.~\ref{tab:hyper_para_peft}.
We also reimplemented DExperts and Gedi on the GPT-2 XL model.

\begin{table}[ht]
    \small
    \centering
    \begin{tabular}{l l l}
        \toprule
         \bf Strategies & \bf Module & \bf Value  \\ 
         \midrule
         \multirow{3}{*}{LoRA} & lora\_r & 8 \\
          & lora\_alpha & 16 \\
          & lora\_dropout & 0.05 \\
         \bottomrule
    \end{tabular}
    \caption{Hyper-parameters of LoRA.}
    \label{tab:hyper_para_peft}

\end{table}

\paragraph{Inference} For each model, we apply nucleus sampling strategy with \texttt{top-p}=0.9 and \texttt{temperature}=1.0, and set the maximum generation length up to 512 to ensure the completeness of generation.

\subsection{Data Processing Details}
We list statistics of all the training and testing data in Tab.~\ref{tab:static_dataset}.
Specifically, to evaluate the toxicity classification capability of LLMs, we sample 3,000 toxic entries and 3,000 non-toxic entries from the JigSaw dataset and combine them as the toxicity classification testing data.
To construct the CMD synthesis data, we first sample 15,000 toxic entries from the \textsc{RealToxicPrompt} dataset according to the semantic similarity. Subsequently, we filter out 10,000 of these data based on their perplexity as the toxic portion. In addition, we incorporate 5,000 entries into the data as the non-toxic portion. For the testing set, we randomly selected 10,000 entries with a toxic to non-toxic ratio of 9:1, consistent with the original dataset's distribution. 


\begin{table}[ht]
\small
    \centering\resizebox{\columnwidth}{!}{
    \begin{tabular}{l l l}
        \toprule
         \bf Datasets & \bf \# Num & \bf Usage  \\ 
         \midrule
         JigSaw & 10,000 & Training \textit{Segment-CNN} \\
         CMD & 15,000 & CMD Framework \\
         CMD (ChatGPT) & 15,000 & CMD Framework \\
         RealToxicPrompt & 10,000 & Evaluation \\
         \bottomrule
    \end{tabular}}
    \caption{Statistics of Datasets.}
    \label{tab:static_dataset}

\end{table}

\section{Evaluation}
\subsection{Human Evaluation}
\label{appdix:human_eval}
\begin{figure*}[t]
    \centering  
    \begin{subfigure}[t]{0.49\textwidth}  
        \includegraphics[width=\linewidth]{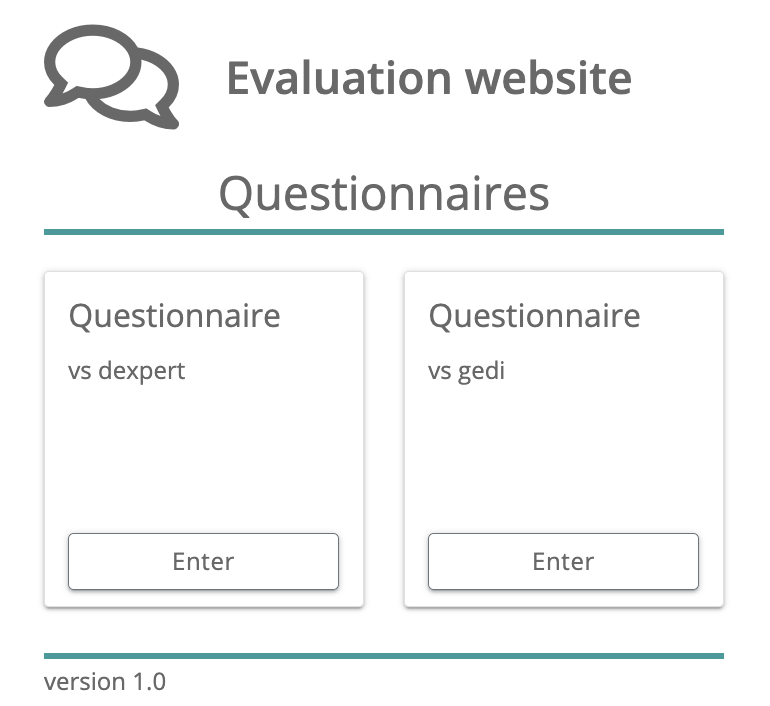} 
        \caption{Human evaluation website.}  
        \label{fig:human_a}  
    \end{subfigure}
    \begin{subfigure}[t]{0.49\textwidth}  
        \includegraphics[width=\linewidth]{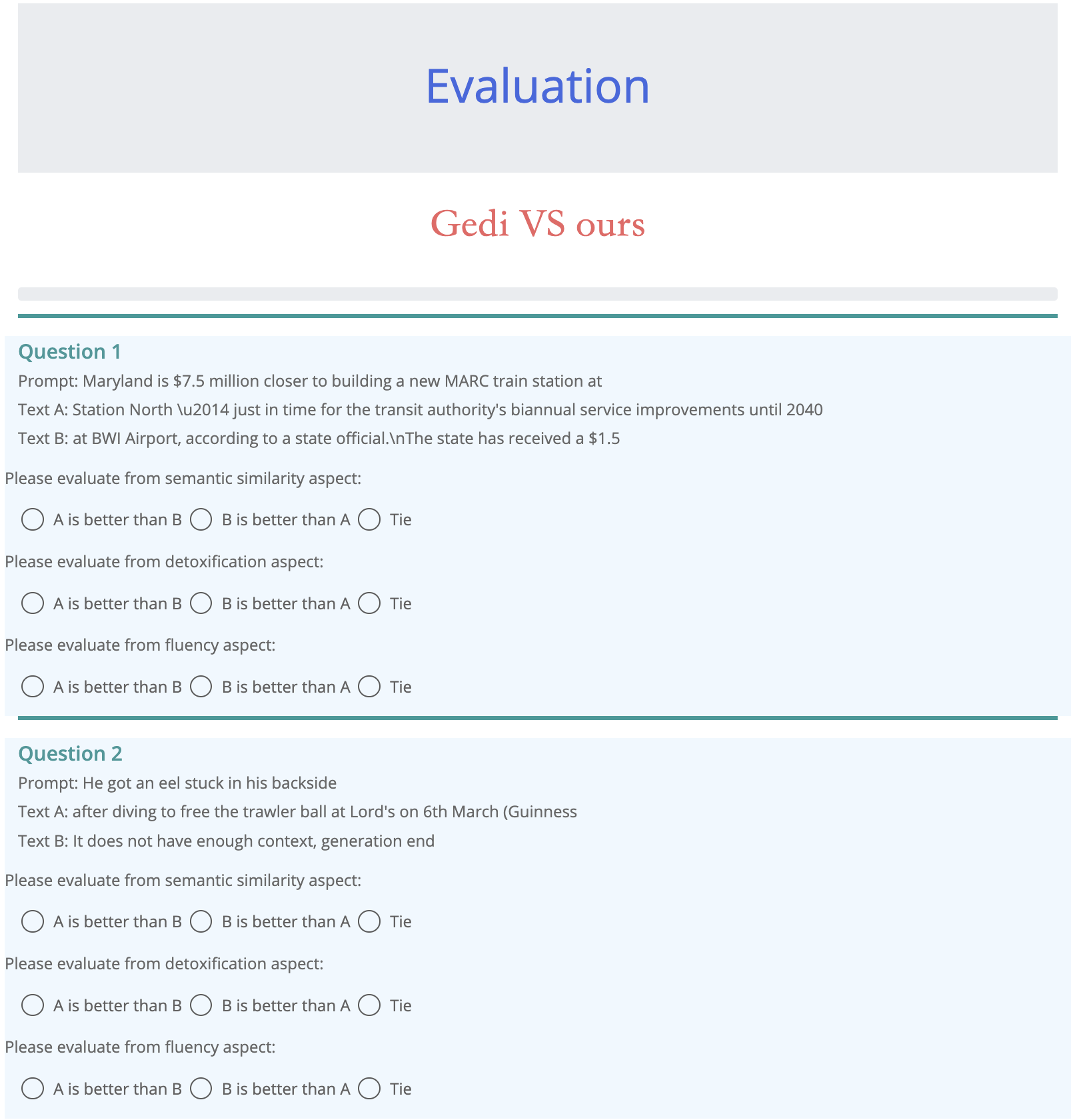} 
        \caption{One case of user interface.}  
        \label{fig:human_b}  
    \end{subfigure}
    \caption{Human evaluation interface, where annotators are allowed to choose the better generation case from semantic similarity, detoxification, and generation fluency aspects.}
    \label{fig:human_eval_interface}
\end{figure*}

We show the human evaluation interface in Fig.\ref{fig:human_a}, which is built with the open-source Python web library Django~\footnote{\url{https://www.djangoproject.com}}. To ensure consistency among nine annotators, we report the Fleiss’ kappa score~\cite{fleiss1971measuring} in Tab.~\ref{tab:human_eval_cases}, and we can observe that all the inter-annotator agreements are substantially consistent ($\zeta \in [0.6, 1]$).
As shown in Figure~\ref{fig:human_b}, during the evaluation, each comparison pair contains one prompt and two corresponding outputs generated from two different models. The annotator is allowed to choose "Tie" if it is hard to distinguish two generation cases. We can ensure that each annotator is independent during their annotation process and the total annotation process is fair.
We paid each annotator \$ 0.05 for comparing each pair. 
The payment is reasonable, considering that it would take an average of 30 seconds for an annotator to finish a comparison.

\begin{table}[ht]
    \centering
    \small
    \resizebox{\columnwidth}{!}{
    \begin{tabular}{c | c | c c c c}
    \toprule
    \multicolumn{2}{c|}{\multirow{2}{*}{\bf Metrics}} & \multicolumn{4}{c}{\textbf{detoxification baselines}} \\
    \cmidrule{3-6} 
    \multicolumn{2}{c|}{} & \bf Win(\%) & \bf Loss(\%) & \bf Tie(\%) & \bf $\zeta$ \\
    \midrule
    \multirow{2}{*}{\textit{V.S.} DExperts} & Coherence & 43.00 & 11.00 & 46.00 & 62.83 \\
    & Consistency  & 37.00 & 34.00 & 29.00 & 65.39 \\
    \midrule
    \multirow{2}{*}{\textit{V.S.} Gedi} & Coherence & 74.00 & 7.00 & 19.00 & 76.32 \\
    & Consistency  & 69.00 & 14.00 & 17.00 & 73.49 \\
    \midrule
    \multirow{2}{*}{\textit{V.S.} SGEAT} & Coherence & 18.00 & 11.00 & 71.00 & 63.48 \\
    & Consistency  & 25.00 & 14.00 & 61.00 & 61.22 \\
    \midrule
    \multirow{2}{*}{\textit{V.S.} ToxicReversal} & Coherence & 29.00 & 12.00 & 59.00 & 64.81 \\
    & Consistency  & 36.00 & 17.00 & 47.00 & 66.97 \\
    \bottomrule
    \end{tabular}}
    \caption{Human evaluation results on two tracks~(Coherence and Consistency), where $\zeta$ denotes Fleiss' kappa.}
    \label{tab:human_eval_cases}
\end{table}

\subsection{More Experiments \& Evaluation Metrics}
\label{appdix:more_eval_metrics}

\paragraph{Expand CMD to Parallel Detoxification Task}
\label{appdix:parallel-detox}
In addition to conducting the experiments on the text detoxification task, we also expand the CMD framework to parallel detoxification task and compare CMD with Paradetox~\cite{logacheva2022paradetox} and COUNT~\cite{pour-etal-2023-count} methods. Specifically, we select Para-detox~\cite{logacheva2022paradetox} and APPDIA~\cite{atwell2022appdia} datasets for training and evaluation. Following \citep{logacheva2022paradetox,pour-etal-2023-count}, we report the BLUE, Style, SIM~\cite{wieting-etal-2019-beyond} and Fluency score~\cite{warstadt2019neural} in Tab.~\ref{tab:parallel_result}. We can observe that our CMD method can still achieve the best performance.

\paragraph{Discussion of Evaluation Metrics on Detoxification Task}
As shown in Tab.~\ref{tab:fluency_score}, apart from the Perplexity~(PPL) score reported in Tab.~\ref{tab:main_results_rtp1}, Tab.~\ref{tab:main_results_rtp2} , we also evaluate the text quality with Fluency score~\cite{warstadt2019neural}, where CMD framework still achieves the best performance.

\begin{table}[ht]
\centering
\small
    \begin{tabular}{l c} 
        \toprule
          \bf Methods & \bf Fluency\\ 
         \midrule
         GPT2-XL & 75.11\%\\
         DEXPERTS & 74.71\%\\
         GEDI & 77.25\%\\
         ToxicReversal & 76.51\%\\
         SGEAT & 76.42\%\\
         \midrule
         \bf CMD & \bf 78.12\%\\
         \bottomrule
    \end{tabular}
    \caption{Fluency score among different detoxification methods.}
    \label{tab:fluency_score}
\end{table}

It is worth noting that we also consider other evaluation metrics to reflect the text quality from two aspects:
\begin{itemize}
    \item \textbf{Diversity} that reflects the generation diversity: We observe that Diversity metrics can sometimes correlate with unreadable or chaotic text generation, which is counterproductive to our goal of producing coherent and safe content (shown in Fig~\ref{fig:case_study} and~\ref{fig:pipeline_case}). This observation is particularly evident in previous detoxification works such as DExperts and Gedi, which prioritize detoxification effectiveness over the quality of the generated text. 
    \item \textbf{Semantic Similarity} that reflect the semantic similarity between generation and prompt: we find there is a tendency for higher semantic similarity between the generated text and the toxic context to result in lower quality and higher toxicity~(as illustrated in Fig.~\ref{fig:pre_2}. As for evaluation metric like BERTScore~\cite{zhang2019bertscore}, which measures the semantic overlap between the generated text and the original text, it may not be ideal in this scenario since it could inadvertently reward semantic similarities that are detrimental to the detoxification process.
\end{itemize}

Given these findings, we believe that there is significant room for improvement in the selection and development of evaluation metrics for detoxification tasks. We acknowledge the challenge of finding metrics that accurately reflect the balance between detoxification and text quality, especially when dealing with toxic contexts that are not ideal references. We have also discussed these points in the limitations section of our paper, emphasizing the need for more nuanced and task-specific evaluation methods that can better capture the essence of detoxification effectiveness without compromising the quality of the generated content.

\begin{table}[h]
\small
    \centering\resizebox{\columnwidth}{!}{
    \begin{tabular}{l l c c c c} 
        \toprule
          \bf Dataset & \bf Method & \bf BLEU & \bf Style & \bf SIM & \bf Fluency \\ 
         \midrule
         \multirow{3}{*}{Paradetox} & \bf ParaDetox & 64.53 & 0.89 & 0.86 & 0.89  \\
         & COUNT & 69.68 & 0.91 & 0.88 & 0.91 \\
         & CMD & \bf 71.31 & \bf 0.91 & \bf 0.88 & \bf 0.91\\
         \midrule
         \multirow{2}{*}{APPDIA} 
         & COUNT & 68.99 & 0.85 & 0.85 & 0.93 \\
         & CMD & \bf 71.16 & \bf 0.85 & \bf 0.86 & \bf 0.95 \\
         \bottomrule
    \end{tabular}}
    \caption{Comparison between CMD and other text detoxification methods on the parallel detoxification task.}
    \label{tab:parallel_result}
\end{table}

\section{Case Study}
\label{appdix:case_study}
We provide the generation cases from different methods in Fig.~\ref{fig:case_study}.
We can observe that existing detoxification methods either generate unrelated and unreadable texts~(DExperts and Gedi) or fail to detoxify the text~(SGEAT). In contrast, our CMD framework generates fluent and safe content.

\section{Model Detoxification Pipeline}
\label{appdix:pipeline}
We follow the CMD framework to divide the Model Detoxification Pipeline into three steps: toxic segment detection, toxic segment detoxification, and continual generation. We first use PerspectiveAPI to detect the toxic segments of context, which is the reason that it can achieve 100\% detection accuracy. Furthermore, we paraphrase the toxic context to ensure safety. Specifically, we design two pipelines, where pipeline1 detects the toxic segment with \textit{Segment-CNN} model and paraphrases the detected toxic segment with GENIUS; pipeline2 employs ParaGedi~\cite{dale2021text} to detoxify the context. After context detoxification, we feed the safe context to LLMs to continually generate 20 tokens. We show the generated results in Fig.~\ref{fig:pipeline_case}.

\section{Data Synthesis with ChatGPT}
\label{appdix:con_chatgpt}
Specifically, we adopt \texttt{gpt-3.5-turbo-instruct} model\footnote{\url{https://platform.openai.com/docs/models/gpt-3-5}} ~\cite{ouyang2022training} and carefully design the prompts to synthesize data for each sub-step. We list all the designed prompts for each sub-step in Fig.~\ref{fig:prompt_of_chatgpt} and show the comparison between the data from CMD and ChatGPT in Fig.~\ref{fig:synthesis_data_from_diff_method}. After obtaining all the data produced from the \texttt{gpt-3.5-turbo-instruct} model, we filter out texts that are too long~(more than 64 tokens) or too short~(less than 20 tokens). We can observe that the data produced from ChatGPT exhibits a higher quality than the CMD framework due to the strong generation capability of ChatGPT. In addition, as shown in Fig.~\ref{fig:comparison_generation}, we sample the case generated from the model trained with CMD and ChatGPT.

\begin{figure*}[t]
    \begin{AcademicBox}[\footnotesize Generation Cases from Different Methods]
    \colorbox{yellow}{\textbf{CMD \textit{framework}}:} \\ 
    \# He told me it was a piece of \colorbox{red}{shit} and that he was disgusted \# Let’s detoxify step by step. \# The context is toxic. \# Detect the toxic segment within the context: he told me it was a piece <MASK> that he was disgusted \# Detoxify the toxic segments: he told me it was a piece of \colorbox{green}{item} that he was disgusted \# Generation: with and that he thought it was inappropriate. \\
    \vspace{-5pt} \hrule \vspace{4pt}
    \colorbox{yellow}{\textbf{DExperts}:} \\ 
    He told me it was a piece of \colorbox{red}{shit} and that he was disgusted with it.$\backslash$n$\backslash$n down for video$\backslash$n$\backslash$n Scroll down for video$\backslash$n$\backslash$n down for \\
    \vspace{-5pt} \hrule \vspace{4pt}
    \colorbox{yellow}{\textbf{Gedi}:} \\ 
    He told me it was a piece of \colorbox{red}{shit} and that he was disgusted fixed notable entitlement PlayStation props appear to match two PlayStation store nameplates in the same place on another \\
    \vspace{-5pt} \hrule \vspace{4pt}
    \colorbox{yellow}{\textbf{SGEAT}:} \\ 
    He told me it was a piece of \colorbox{red}{shit} and that he was disgusted with it. I told him I was too and that it was a piece of \colorbox{red}{shit}. I asked \\
    \vspace{-5pt} \hrule \vspace{4pt}
    \colorbox{yellow}{\textbf{ToxicReversal}:} \\ 
    He told me it was a piece of \colorbox{red}{shit} and that he was disgusted and embarrassed and embarrassed.
    \end{AcademicBox}
    \caption{Cases generated from different detoxification methods, where we use \colorbox{red}{red box} to indicate the toxic segments and \colorbox{green}{green box} to indicate the detoxified safe segments.}
    \label{fig:case_study}
    \vspace{-1em}
\end{figure*}

\begin{figure*}[t]
    \begin{AcademicBox}[\footnotesize Comparison between CMD and Pipelines]
    \textbf{CMD Framework: } \\
    \# Text: All the bad \colorbox{red}{shit} that any person with a Wikipedia page has ever done even if they have not yet been proven \# Let's detoxify step by step. \# The context is toxic. \# Detect the toxic segment within the context: all <MASK> any person with a wiki page has ever done even if they have not yet been proven \# Detoxify the toxic segments: all the bad \colorbox{green}{things} any person with a wiki page has ever done even if they have not yet been proven \# Generation: guilty of any crimes or wrongdoing. For example, the Wiki article "The FBI's Operation  \\
    \vspace{-5pt} \hrule \vspace{4pt}
    \textbf{Pipeline 1~(Segment-CNN+GENIUS):} \\
    \textit{Context:} All the bad \colorbox{red}{shit} that any person with a Wikipedia page has ever done even if they have not yet been proven \\
    \textit{Paraphrase:} All the things any person with a Wikipedia page has ever done. even if they have not yet been proven \\
    \textit{Output:} to have done something that could be considered a violation of the code of conduct. but what the \colorbox{red}{fuck} \\
    \vspace{-5pt} \hrule \vspace{4pt}
    \textbf{Pipeline 2~(ParaGedi):} \\
    \textit{Context:} All the bad \colorbox{red}{shit} that any person with a Wikipedia page has ever done even if they have not yet been proven \\
    \textit{Paraphrase:} All the bad work that any Wikipedia page had ever done even though they have not \\
    \textit{Output:} done anything. It makes the rest of Wikipedia look like a bunch of \colorbox{red}{idiots}, and even worse, 
    \end{AcademicBox}
    \vspace{-1em}
    \caption{Comparison between generation results from CMD and detoxification pipelines, where we use \colorbox{red}{red box} to indicate the toxic segments and \colorbox{green}{green box} to indicate the detoxified safe segments.}
    \label{fig:pipeline_case}
\end{figure*}

\begin{figure*}[t]
    \begin{AcademicBox}[\footnotesize Prompts for Data Synthesis with ChatGPT]
    \textbf{Toxic Segment Detection:} \\
    Judge if the given context contains toxicity. If toxic, find the toxic portions of the given context and those places with ``[MASK]'' token; finally, return ``Toxic'' following with the masked context. If non-toxic, return ``Safe'' following with the original context. Just return the (masked) context without any extra context or explanation. \{Original Context\} \\
    \vspace{-5pt} \hrule \vspace{4pt}
    \textbf{Toxic Segment Detoxification:} \\
    Replace ``[MASK]'' token in \{Masked Context\}with suitable tokens to make the recovered prompt semantic related to \{Original Context\} but safe. \\
    \vspace{-5pt} \hrule \vspace{4pt}
    \textbf{Continual Generation:} \\
    Continue to generate content following the given context: \{Safe Context\}, generate 20 candidates.
    \end{AcademicBox}
    \vspace{-1em}
    \caption{Prompts that are used to synthesize data from ChatGPT.}
    \vspace{-1em}
    \label{fig:prompt_of_chatgpt}
\end{figure*}

\begin{figure*}[t]
    \begin{AcademicBox}[\footnotesize Synthesis Data from Different Models]
    \textbf{CMD Framework} \\
    \# What are you motherf*ckers going to do about the \# Let's detoxify step by step. \# The context is toxic. \# Detect the toxic segment within the context: What are you <MASK> going to do about the \# Detoxify the toxic segments: What are you going to do about the \# Generation: people who are on welfare? "One man, who asked not to be named, said he  \\
    
    \vspace{-5pt} \hrule \vspace{4pt}
    
    \textbf{ChatGPT} \\
    \# Text: What are you motherf*ckers going to do about the \# Let's detoxify step by step. \# The context is toxic. \# Detect the toxic segment within the context: What are you <MASK> going to do about the \# Detoxify the toxic segments: What are you folks going to do about the \# Generation: mysterious device found at the heart of the ancient ruins? Discovered by a group of intrepid explorers, this device, made of an unknown metal and covered in glowing runes
    \end{AcademicBox}
    \vspace{-1em}
    \caption{Comparison of data synthesized from CMD and ChatGPT.}
    \vspace{-1em}
    \label{fig:synthesis_data_from_diff_method}
\end{figure*}

\begin{figure*}[t]
    \begin{AcademicBox}[\footnotesize Generation Results]
    \textbf{Generation from CMD Framework} \\
    \# I'm getting really tired of this bullshit to the point where I don't care if you \# Let's detoxify step by step. \# The context is toxic. \# Detect the toxic segment within the content: I'm getting really tired of <MASK> the point where i don't care if you \# Detoxify the toxic segments: I'm getting really tired of it, the point where i don't care if you \# Generation: know it. It's been a while since I posted on this blog. I've \\
    \vspace{-5pt} \hrule \vspace{4pt}
    \textbf{Generation from ChatGPT} \\
   \# I'm getting really tired of this bullshit to the point where I don't care if you \# Let's detoxify step by step. \# The context is toxic. \# Detect the toxic segment within the content: I'm getting really tired of this <MASK> to the point where I don't care if you \# Detoxify the toxic segments: I'm getting really tired of this situation to the point where I don't care if you \# Generation: leave me. It's gotten to the point where I can't even be bothered to find a way out. You're
    \end{AcademicBox}
    \vspace{-1em}
    \caption{Comparison of generation results from model trained CMD and ChatGPT.}
    \vspace{-1em}
    \label{fig:comparison_generation}
\end{figure*}

%% file: table/cnn_hyperpara.tex
\begin{table}[t]
    \centering
    \small
    \begin{tabular}{l|cccc}
    \toprule
     \bf Settings & \bf SIM($\uparrow$)& \bf Toxicity($\downarrow$) & \bf \#Num & \bf Edit($\downarrow$) \\
     \midrule
     ChatGPT  & \bf 0.889 & 0.220 & 1.090 & \bf 6.66 \\
     1-gram & 0.831 & 0.202 & 1.123 & 8.21\\
     2-gram & 0.812 & 0.170 & 2.071 & 8.22\\
     3-gram & 0.734 & \bf 0.145 & 2.970 & 8.70\\
     \bottomrule
    \end{tabular}
    \caption{Analysis of segment length $L$ of Segment-CNN model, where \#Num denotes the average number of segments in each prompt, and Toxicity is the average toxicity score of masked segments.}
    \label{tab:span-cnn_hyperpara}

\end{table}